\documentclass[shortAfour,times,sageh]{sagej}

\usepackage{moreverb,url}
\usepackage[colorlinks,bookmarksopen,bookmarksnumbered,citecolor=red,urlcolor=red]{hyperref}

\newcommand\BibTeX{{\rmfamily B\kern-.05em \textsc{i\kern-.025em b}\kern-.08em
T\kern-.1667em\lower.7ex\hbox{E}\kern-.125emX}}

\usepackage{amsthm,amssymb,natbib,url,amsmath}
\usepackage{graphicx}
\usepackage{graphics}
\usepackage{clrscode3e}
\usepackage{caption}
\usepackage{subcaption}
\usepackage{bm}
\usepackage[author={Caelan Garrett}, color=blue, icon=note]{pdfcomment}

\theoremstyle{plain}\newtheorem{thm}{Theorem}
\theoremstyle{definition}\newtheorem{defn}{Definition}
\theoremstyle{plain}\newtheorem{lem}{Lemma}
\theoremstyle{plain}\newtheorem{cor}{Corollary}

\usepackage{xcolor,colortbl}
\definecolor{Gray}{gray}{0.85}
\newcolumntype{g}{>{\columncolor{Gray}}c}
\newcolumntype{w}{>{\columncolor{white}}c}

\newcommand{\prob}{PPM}
\newcommand{\ffrob}{{\sc FFRob}}
\newcommand{\crg}{{\sc CRG}}

\def\volumeyear{2016}

\begin{document}

\runninghead{Garrett et al.}

\title{FFRob: Leveraging Symbolic Planning for Efficient Task and Motion Planning}

\author{Caelan Reed Garrett\affilnum{1}, Tom\'as Lozano-P\'erez\affilnum{1}, and Leslie Pack Kaelbling\affilnum{1}}

\affiliation{\affilnum{1}MIT CSAIL, USA}

\corrauth{Caelan Reed Garrett,
Computer Science and Artificial Intelligence Laboratory,
32 Vassar Street,
Cambridge, MA 02139 USA}

\email{caelan@csail.mit.edu}

\begin{abstract}
Mobile manipulation problems involving many objects are challenging to solve due to the high dimensionality and multi-modality of their hybrid configuration spaces. 
Planners that perform a purely geometric search are prohibitively slow for solving these problems because they are unable to factor the configuration space. 
Symbolic task planners can efficiently construct plans involving many variables but cannot represent the geometric and kinematic constraints required in manipulation. 
We present the \ffrob{} algorithm for solving task and motion planning problems. First, we introduce Extended Action Specification (EAS) as a general purpose planning representation that supports arbitrary predicates as conditions. 
We adapt existing heuristic search ideas for solving \proc{strips} planning problems, particularly delete-relaxations, to solve EAS problem instances. 
We then apply the EAS representation and planners to manipulation problems resulting in \ffrob{}.
 \ffrob{} iteratively discretizes task and motion planning problems using batch sampling of manipulation primitives and a multi-query roadmap structure that can be conditionalized to evaluate reachability under different placements of movable objects. 
 This structure enables the EAS planner to efficiently compute heuristics that incorporate geometric and kinematic planning constraints to give a tight estimate of the distance to the goal. 
Additionally, we show \ffrob{} is probabilistically complete and has finite expected runtime. Finally, we empirically demonstrate \ffrob{}'s effectiveness on complex and diverse task and motion planning tasks including rearrangement planning and navigation among movable objects.
\end{abstract}


\maketitle



\section{Introduction}

A long-standing goal in robotics is to develop robots that can operate autonomously in unstructured human environments. 
Recent hardware innovations have made mobile manipulator robots increasingly affordable, and sensing innovations provide unprecedented sensory bandwidth and accuracy.
Progress in algorithms for navigation and motion planning has enabled some basic forms of mobile manipulation, which combine actuation of a robot's base and end-effectors to move objects in the world.
However, mobile manipulation is primarily restricted to picking and placing objects on relatively uncluttered surfaces. 
Planning for mobile manipulation problems involving cluttered environments and multiple manipulation primitives still presents substantial challenges.

\begin{figure}[h]
\centering
\includegraphics[width=0.40\textwidth]{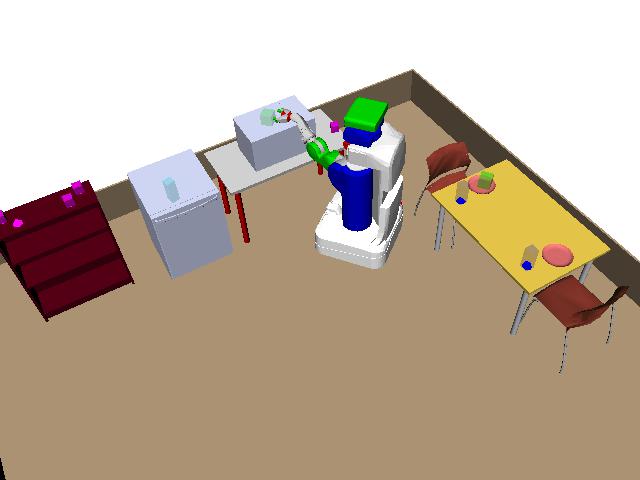}
\caption{A task and motion planning problem requiring cooking dinner. The robot must obtain two green cabbages from the shelves, clean them on the dishwasher, cook them on the microwave, and serve them. Additionally, the robot must organize the dirty cups on the table, clean them, and set the table.} 
\label{fig:wall-namo}
\end{figure}

Researchers in artificial intelligence planning~\citep{Ghallab04} have been tackling
problems that require long sequences of actions and large discrete state-spaces.
However, these symbolic ``task-level'' planners do not naturally
encompass the detailed geometric and kinematic considerations that
robot motion planning requires. The original Shakey and \proc{strips} robot
system~\cite{Fikes71,Nilsson84}, from which many of these symbolic
planners evolved, managed to plan for an actual robot by working in a
domain where all legal symbolic plans were effectively executable.
This required the ability to represent symbolically a sufficient
set of conditions to guarantee the success of the steps in the plan.
Compactly encoding success conditions using typical symbolic representations is not generally possible in realistic manipulation
domains because the geometrical and kinematic constraints are significant.

Consider a simple manipulation domain where a variety of
objects are placed on a table and the robot's task is to collect some
subset of the objects and pack them in a box.
The basic robot operations are to pick up an object and place it somewhere else; in
addition, the robot can move its base in order to reach a distant
object.  Note that, in general, to pick a distant object or place an object at a distant location, the robot will have to
move other objects out of the way.  Which objects need moving depends
on their shapes, the shape of the robot, where the robot's base is placed and
what path it follows to the object.  When an object is moved, the
choice of where to place it requires similar considerations.  The key
observation is that constructing a valid symbolic plan requires access
to a characterization of the connectivity of the underlying free
configuration space (for the robot and all the movable objects).  We
cannot efficiently maintain a representation of this connectivity with a set of static
assertions updated by symbolic actions; determining how the
connectivity of the underlying free space changes requires geometric
computation.

Whereas classic robot motion planning requires a search
in the robot configuration space, manipulation planning requires a
search in the combined configuration space of the robot and all the
movable objects in the world. Achieving a manipulation goal requires
choosing which object to move when, which grasps and intermediate
placements to use, {\it etc}. 

Manipulation planning remains challenging because it is notoriously difficult to work in a high-dimensional space and
make a long sequence of intertwined decisions.
Existing manipulation planning algorithms~\citep{simeon2004manipulation,Cambon,Hauser,HauserIJRR11,Dogar12,barry2013hierarchical} take substantial time to plan operations involving relatively few objects. 
Without any search guidance, these algorithms must explore a large fraction of the configuration space to find a plan. 
And the size of the configuration space grows exponentially in the number of moveable objects in the world.
Constructing such plans generally requires some methods for partitioning the problem and for effective search guidance.
Therefore we seek to integrate the capabilities of a task planner and a manipulation planner to produce an efficient mobile manipulation planning algorithm.

\subsection{Approach}

The primary contribution of this paper is \ffrob{}, an efficient and probabilistically complete algorithm for fully integrated task and motion planning. 
This paper is an extended and revised version of a conference paper by~\cite{GarrettWAFR14}. 
We model task and motion planning as symbolic planning where the conditions of actions are complex predicates involving geometric and kinematic constraints. 
We adapt efficient existing heuristic search algorithms for solving traditional symbolic planning problems to solve task and motion planning problems. 
The key computational benefit of the approach is that it is able to incorporate geometric and kinematic constraints in the heuristic to strongly guide the search. 

To start, we formally identify a subclass of task and motion planning problems, pick-place-move (\prob{}) problems, that will be our focus.
We later show how \ffrob{} can be easily extended to solve more general task and motion planning problems involving additional symbolic inferences or manipulation primitives.

We introduce Extended Action Specification (EAS), a new symbolic planing representation that supports complex conditions. 
Although this representation is not specific to task and motion planning or even robotics problems, our primary application of it will be to \prob{} problems discretized using sampling. 
EAS is able to represent actions with complex conditions much more concisely than a traditional symbolic planning representation. 
Additionally, EAS allows specification of predicate evaluation functions to efficiently test conditions. 
In the context of \prob{} problems, we give a method for quickly evaluating reachability predicates using dynamic programming and collision check caching. 
Following this, we give our extension of relaxed planning heuristics, particularly the FastForward (FF) heuristic~\cite{HoffmannN01}, to the EAS planning representation. 

In order to frame task and motion planning as symbolic planning in a finite domain, we repeatedly discretize the planning problem. 
This involves batch sampling a set of placement poses and grasp transforms to identify the pick and place actions. 
Then, we construct a roadmap of robot configurations to give an approximation of the robot's free configuration space. 
This roadmap is instrumental in enabling efficient evaluation of reachability predicates that arise when the robot seeks to move to a new configuration.

We prove completeness results for \ffrob{} by identifying a class of non-degenerate \prob{} problems and proving \ffrob{} will solve them with finite expected runtime. 
Finally, we perform experiments on challenging manipulation problems and explore the effect of various planner configurations on their performance. 
We demonstrate that \ffrob{} can solve a broad class of feasible task and motion planning problems that involve navigating among and rearranging moveable objects. 


\section{Related Work}

This work draws from existing approaches to manipulation planning and to task and motion planning
well as ideas from the artificial intelligence symbolic planning literature.
Our focus will be on showing how ideas originally developed for symbolic planning can be adapted to continuous-action domains to more efficiently solve high-dimensional task and motion planning problems.

\subsection{Manipulation Planning}

In manipulation planning, the goal is not just to move the robot
without collision, as in classical motion planning, but also to operate on the objects in the world.
This problem was addressed from the earliest days of algorithmic
motion planning, for example by~\cite{LozanoPerez81},~\cite{handeyICRA87}, and~\cite{Wilfong89}.  The modern treatment of this problem, involving continuous grasps as well as continuous placements, was
pioneered by~\cite{Alami91, AlamiTwoProbs} who introduced
the {\em manipulation graph}.  This graph breaks the problem of one
robot moving one object in a potentially complex environment into
several problems of moving between connected components of the
combined configuration space where each component shares the same grasp.  A solution
is an alternating sequence of transit paths, in which the robot is not
grasping an object, and transfer paths, in which it is.
~\cite{simeon2004manipulation} expanded this work to more realistic
settings by using probabilistic roadmaps. They looked
at manipulations necessitating multiple regrasps. Their approach uses
the manipulation graph to identify a high-level sequence of transit
and transfer paths then performs the motion planning required to
achieve them. 

~\cite{StilmanWAFR06} and ~\cite{StilmanICRA07} address a version of
manipulation planning called {\em navigation among movable obstacles}
(NAMO), where the robot must reach a specified location among a field
of movable obstacles. In order to solve monotonic NAMO instances, 
instances requiring at most one pick and place for each object,
they plan backwards from the goal and use swept volumes to determine, recursively, which additional objects must be moved.
~\cite{van2009path} developed a
probabilistically complete algorithm for NAMO.  However, this
algorithm assumes that one can fully characterize the connected
components of the configuration space of the robot at each planning
step; this is computationally prohibitive for robotic configuration spaces
with more than two dimensions.

~\cite{Hauser} and ~\cite{HauserIJRR11} identified a generalization of
manipulation planning as {\em hybrid planning}, that is, planning for
systems with multiple (possibly infinitely many) {\em modes},
representing different constraint sub-manifolds of the configuration
space.
In a robotics domain, for example, modes are characterized by a grasp on a particular object and the placements of movable objects. 
The key insight is that, as in the manipulation graph, one can conceptualize the planning process as
alternating between moving in a single mode, where the constraints are
constant (e.g., moving through free space with a grasped object), and
switching between modes (e.g. grasping a new object).  So, solving
these problems requires being able to plan within a single mode 
and identifying configurations where modes can change, which is generally specific to
the task.  Hauser provided a probabilistically complete algorithm that solves problems of this type assuming that effective
single-mode planners and mode-transition samplers are available.
However, a pure uni-directional sampling-based algorithm has trouble
solving high-dimensional problems.  

\cite{barry2013hierarchical} defined a bidirectional rapidly-exploring random tree (RRT) search
of the combined configuration space. Importantly, the individual moves of this
algorithm consider complete plans to reach the goal, ignoring
obstacles, ensuring that suggested motions have some chance of
being on the path to the goal.  They also investigated a two-layer
hierarchy, where the higher level plans only for the manipulated
object (without the robot), with the objective of identifying relevant
mode transitions to guide the full planner.  
This planner was limited to domains with one movable object and had running times on the order of minutes.

\cite{krontirisRSS2015,krontiris2016icra} provided an algorithm for rearrangement planning: a special instance of pick-and-place planning where all objects have explicit goal poses. Their method constructs a Probabilistic Roadmap (PRM)~\citep{Kavraki98probabilisticroadmaps} in the combined configuration space similar to that of~\cite{barry2013hierarchical}. It samples arrangements of the objects and uses a greedy local planner based on the NAMO algorithm of~\cite{StilmanWAFR06} to connect an existing PRM vertex with the sampled object arrangement. The use of the PRM was able to recover completeness for problems that could not be solved by just the greedy planner. However, the lack of search guidance forces the planner to explore a large number of object arrangements.

\cite{GarrettIROS15} introduced the Hybrid Backward-Forward (HBF) algorithm for 
hybrid planning problems. HBF uses a backward search to produce successors and distance estimates for states in
a forward search. HBF was applied to manipulation problems involving robot
primitives for picking, placing, and pushing. In contrast with \ffrob{}'s batch action sampling, HBF
samples action primitives while simultaneously searching through the state-space. 

\cite{King2016} investigated rearrangement planning with both object-centric motions, actions involving a particular object, and robot-centric motions, actions not involving any particular objects. Most of the presented literature involves only object-centric motions. Planning with robot-centric motions can enable complex manipulations such as multi-object pushing and whole arm manipulation. Using these primitives can result in much shorter and more natural plans than using object-centric motions alone. Like~\cite{barry2013hierarchical}, they generalize the RRT algorithm to plan with both of these motions. Their algorithm also lacks strong search guidance needed to effectively plan for problems with many objects and long horizons.

\subsection{Symbolic Planning}

The artificial intelligence (AI) planning community has largely adopted heuristic state-space search
methods for solving symbolic planning problems. These planning problems are, by and large, discrete problems
that are represented using Planning Domain Definition Language (PDDL) as proposed by~\cite{mcdermott1998pddl}.

~\cite{BonetG99,bonet2001planning} popularized these state-space search methods by showing how {\em domain-independent} heuristics could be derived automatically, by manipulating the conditions and effects of the actions. 
The key idea is to define a relaxed version of the planning problem where the
actions do not have any ``delete'' effects, that is, no previously
achieved result becomes false.  This is an easier problem to solve
than the original problem and the solution can be used to provide an
approximation to the actual cost of solving a problem.  They identified
two heuristics, $h_{add}$ and $h_{max}$, that derive their estimates
from a relaxed version of the {\em plan graph}~\cite{blum1997fast}.
They each can be computed in polynomial time by taking the sum or max of the costs of achieving
individual terms in a conjunctive goal respectively.  The $h_{max}$ heuristic is admissible,
while $h_{add}$ is inadmissible but more informed (it tends to be closer to the true cost in
practice, providing more effective guidance).

The FastForward (FF) planning system of~\cite{HoffmannN01} introduced the $h_{\it ff}$ heuristic, which
explicitly extracts a plan from the relaxed plan graph and uses the plan's cost as its value.
By avoiding double-counting actions that achieve several conditions, $h_{\it ff}$ is generally a tighter estimate of the cost to reach the goal than $h_{add}$ and $h_{max}$.
Importantly, the resulting relaxed plans can also be used to suggest useful actions to consider at the current state, which can reduce the branching factor of the search.

The AI planning community has also investigated planning in hybrid domains with simple continuous dynamics.
\cite{coles2013hybrid} gave a heuristic for numerical planning that
combines a mixed integer program with a relaxed plan graph to create a hybrid
heuristic that is able to more strongly guide the search by using fewer approximations.
This adaptation of a relaxed plan graph, albeit in a different way and for a different problem,
is similar in spirit to the inclusion of geometric inferences in \ffrob{}'s relaxed plan graph.

More generally, Planning Modulo Theories (PMT)~\cite{gregory2012planning} is a framework for using arbitrary first-order logical theories in planning problems. This formulation, inspired by SAT Modulo Theories (SMT), was designed to have wide expressivity and unify the representation for many existing planning types. Gregory et al. also gave a heuristic search algorithm for solving PMT problems. Its heuristic is a extension of $h_{max}$. 
The resulting planner is able to solve many problems that cannot be modeled with PDDL.
It even outperforms some algorithms operating on PDDL problems because it plans using a more compact representation by allowing complex conditions. 
Our planning representation and algorithms are similar to PMT applied to problems with arbitrary propositional conditions. However, our framework allows for custom evaluation of complex conditions and supports additional heuristics that are more effective than $h_{max}$.

\subsection{Task and Motion Planning}

There have been a number of approaches to integrating discrete task
planning and continuous motion planning in recent years. 
The pioneering Asymov system~\citep{Cambon} conducts an interleaved search at the symbolic and
geometric levels. They carefully consider the consequences of using
non-terminating probabilistic algorithms for the geometric planning,
allocating computation time among the multiple
geometric planning problems that are generated by the symbolic
planner.  The process can be viewed as using the task planner as a heuristic to guide
the motion planning search. However, since the task-level
planner is ignoring geometry, its value as a heuristic is quite
limited. The work of~\cite{Plaku}
is similar in approach. 

A natural extension to the classic symbolic planning paradigm is to
introduce ``computed predicates'' (also known as ``semantic
attachments''); that is, predicates whose truth value is established
not via assertion but by calling an external program that operates on
a geometric representation of the state~\cite{dornhege09icaps,dornhege13irosws}.
A motion planner can serve to implement such a predicate, determining the reachability of one
configuration from another.  
A difficulty with this approach, however, is that calling a motion planner is generally expensive. 
This leads to a desire to minimize the set of object placements considered to limit the branching factor of the search.
Considering only a sparse set of placements may limit the generality of the planner.
Additionally, computed predicates are ignored during heuristic computation.  
This leads to a heuristic that is uninformed about geometric considerations and may result in considerable inefficiency due to heuristic plateaus. 
The work of~\cite{Erdem}, is similar in approach to~\cite{dornhege09icaps}, augmenting a task planner that is based on explicit causal reasoning with the ability to check for the existence of paths for the robot. 

\cite{LagriffoulDSK12,lagriffoul2014efficiently} interleave the symbolic and geometric searches and focus on limiting the amount of geometric backtracking.
They generate a set of approximate linear constraints imposed by the program under consideration, e.g.,
from grasp and placement choices, and use linear programming to
compute a valid assignment or determine that one does not exist.  This
method is particularly successful in domains such as stacking objects
in which constraints from many steps of the plan affect geometric choices.
Although their approach is able to efficiently decide if a task-level plan is geometrically feasible, 
it is unable to inform the task-level search which may result in attempting many infeasible plans.

\cite{Pandey12} and~\cite{deSilva} use
HTNs instead of generative task planning. Their system can backtrack over choices made by the geometric
module, allowing more freedom to the geometric planning than in the
approach of~\cite{dornhege09icaps}. 
In addition, they use a cascaded approach to computing difficult applicability conditions:
they first test quick-to-evaluate approximations of accessibility 
predicates, so that the planning is only attempted in situations in
which it might plausibly succeed.

In the HPN approach of~\cite{HPN},
a regression-based symbolic planner uses {\em generators}, which
perform fast approximate motion planning, to select geometric
parameters, such as configurations and paths, for the actions.
Reasoning backward using regression allows the goal to significantly
bias the actions that are considered.

\cite{Srivastava14} offer a
novel control structure that avoids computing expensive condition
values in many cases by assuming a favorable default valuation of the
condition elements; if those default valuations prove to be
erroneous, then it is discovered in the process of performing
geometric planning to instantiate the associated geometric action.
In that case, symbolic planning is repeated after adding updated valuations.  This approach requires
the ability to diagnose why a motion plan is not possible in a given
state, which can be challenging, in general.

\cite{lozano2014constraint} leverage constraint satisfaction problem (CSP) solvers for task and motion planning.
Their approach performs a discrete search in the space of plan skeletons and uses a CSP solver
to determine if a valid set of action parameters completes the plan skeleton.
\cite{dantam2016tmp} extend this approach by more generally formulating task and motion planning as a satisfiability modulo theories (SMT) problem. They use an incremental constraint solver to add motion constraints to the task-level logical formula when a candidate task plan is found. 
Upon failure, they iteratively increase the plan depth and motion planning timeouts, which results in a probabilistically complete algorithm.

\cite{toussaint2015logic} formulates task and motion planning as a logic-geometric program, a non-linear constrained optimization problem augmented with a logic and knowledge base.
He introduces three approximations that make solving the problem more tractable by sequentially optimizing the final state, transfer configurations, and motion trajectories.
His experiments apply the technique to maximizing the height of a stable structure constructed from a set of objects. 

All of these approaches, although they have varying degrees of integration of the symbolic and geometric planning, generally lack a true integrated search that allows the geometric details to affect the focus of the symbolic planning. \ffrob{} develops such an integrated search, provides methods for performing it efficiently, and shows that it results in significant computational savings.


\section{Problem Formulation} 
\label{sec:formulation}

We start by modeling robotic planning domains that involve a single manipulator on a mobile base in an environment with moveable rigid objects. 
We focus on this specific domain because it is the subject of our experiments; however, the general formulation has broader applicability and can be extended to different domains involving, for instance, several manipulators or additional symbolic fluents. 
We call this class of problems {\it pick-place-move} (\prob{}) problems. 
We assume that the environment is fully observable and that actions have deterministic effects. 

\begin{defn}
A {\em \prob{} domain} ${\cal D} = \langle {\cal Q}, \{({\cal P}^{o_1}, {\cal G}^{o_1})..., ({\cal P}^{o_m}, {\cal G}^{o_m})\} \rangle$ is specified by a robot configuration space ${\cal Q}$ as well as a space of placement surfaces ${\cal P}^{o_i}$ and a space of grasps ${\cal G}^{o_i}$ for each of the $m$ moveable objects $o_i$. 
\end{defn}


${\cal P}^{o_i}$ is the union of poses where object $o_i$ can legally be placed such as poses supported by tops of tables or floors. ${\cal G}^{o_i}$ contains a set of grasps which may be discrete or continuously infinite depending on the geometry of the robot and $o_i$. We assume that ${\cal Q}$ and each ${\cal P}^{o_i}$ take into account collisions with any fixed obstacles or joint limits, so values in each space are collision-free when no moveable objects are in the environment. We will not consider stacking domains where ${\cal P}^{o_i}$ could contain surfaces on top of other objects. \ffrob{} can be extended to solve stacking problems by sampling sets of object poses that form a structurally sound stack. 
This formulation encompasses pick-and-place planning, rearrangement planning, and navigating among moveable objects (NAMO). We will later show that additional symbolic values can be easily incorporated into the domain to plan for tasks like cooking meals.

In a \prob{} domain ${\cal D}$, we can represent the state of the system using a set of {\em variables} ${\cal V} = \{v_r, v_h, v_{o_1}, ..., v_{o_m}\}$. Each variable $v_a$ has a domain of possible values $D_a$. A {\em state} $s = \{v_r=q, v_h=g, v_{o_1}=p^{o_1}, ..., v_{o_m}=p^{o_m} \}$ is an assignment of values to the variables. These variables along with their domains and values are as follows:
\begin{itemize}
\item $v_r$ is the robot configuration variable. The robot configuration domain is just $D_r = {\cal Q}$. Each configuration $q \in D_r$ specifies the pose of the base as well as the joint angles of the manipulator.
\item $v_h$ is the robot holding variable. The robot holding domain is $D_h = {\cal G}^{o_1} \cup ... \cup {\cal G}^{o_m} \cup \{\kw{None}\}$. For $g \in D_h$, $g = \kw{None}$ indicates the robot is not holding anything. Otherwise, $g = (o, \gamma)$ indicates the robot is holding object $o$ with a grasp transform $\gamma$ relating robot's end-effector pose and the object pose.
\item $v_{o_i}$ is the object $o_i$ pose variable for object label $i \in (1, ..., m)$. The object pose domain is $D_{o_i} = {\cal P}^{o_i} \cup \{\kw{None}\}$. For $p^{o_i} \in D_{o_i}$, $p^{o_i} =  \kw{None}$ indicates that object $o_i$ is not placed. Otherwise, $p^{o_i} = (x, y, z, \theta)$ is a four-dimensional pose (we assume that the object is resting on a stable face on a static horizontal surface).
\end{itemize}

We assume that the world is quasi-static and the robot can only hold a single object. When the robot is holding an object $o_i$, the object pose can be determined using $p^{o_i} = q \times \proc{transform}(g)$ where $\proc{transform}(g) = \gamma$. As such, it is redundant to explicitly update the pose of $o_i$ when it is in the hand, so we let $p^{o_i} = \kw{None}$ for simplicity. A state is {\em legal} if there are no collisions among the robot, held object, and the placed objects. 

\begin{defn}
A {\em \prob{} problem} $\Pi = \langle s_0, S_*\rangle$ in a \prob{} domain ${\cal D}$ is specified by an initial state $s_0$ and a set of goal states $S_*$.
\end{defn}

The initial state $s_0 = \{v_r=q_0, v_h=g_0^{o_{h_0}}, v_{o_1}=p_0^{o_1}, ..., v_{o_m}=p_0^{o_m} \}$ must be a legal state of the system. For simplicity, we will assume that $S_*$ can be represented as the conjunction of goal sets for individual variables rather than logical predicates. We make this restriction because our manipulation experiments only involve goals that can be expressed in this form, and introducing arbitrary goal predicates will complicate the theoretical analysis. Thus, $S_* = \{v_r \in Q_*, v_h \in G_*^{o_{h_*}}, v_{o_1} \in P_*^{o_1}, ..., v_{o_m} \in P_*^{o_m} \}$ defines a set of legal states in the Cartesian product of $Q_* \times G_*^{o_{h_*}} \times P_*^{o_1} \times ... \times  P_*^{o_m}$ where $Q_* \subseteq D_r$, $G_*^{o_{h_*}} \subseteq D_h$, and $P_*^{o_i} \subseteq D_{o_i}$ for $i \in [m]$. If the goal set is left unspecified for a variable, the goal set defaults to the full variable domain.  

\section{The FFRob Algorithm Overview}

%

At the highest level of abstraction, \ffrob{} iteratively alternates between a sampling phase and a planning phase until it is able to find a solution. The sampling phase discretizes the \prob{} problem by creating symbolic actions from a finite sampled set of poses, grasps, and configurations. The planning phase performs a discrete search to decide whether a solution exists. If the discrete search fails to find a solution, the process repeats with a larger set of samples. 

The pseudocode for \ffrob{} is presented in figure~\ref{fig:ffrob}. \ffrob{}'s inputs are a \prob{} domain ${\cal D}$ and problem $\Pi$, and its output is a solution \id{plan}. The procedure begins by initializing a set of sampling parameters $\theta$ that govern the number of samples to produce using $\proc{initial-parameters}$. 
In the sampling phase, \proc{sample-discretization} (figure~\ref{fig:discretization}) discretizes the \prob{} problem by sampling a specified number of configurations, poses, and grasps determined by $\theta$. \proc{sample-discretization} returns a symbolic planning representation of the goal $C_*$ and actions $A$ in the current discretization of the problem.
In the planning phase, \proc{search} (figure~\ref{fig:search}) performs a discrete search using $C_*$ and $A$. \ffrob{} immediately terminates if it finds a solution. Otherwise, $\theta$ is increased using \proc{increment-parameters}, and this process repeats.

The majority of this paper is dedicated to the implementation of the two key subroutines: \proc{sample-discretization} and \proc{search}. Section~\ref{sec:rob} discusses the discretization created by \proc{sample-discretization}.
Sections~\ref{sec:rep}, \ref{sec:search}, and \ref{sec:heuristic} are concerned efficient search algorithms that  implement \proc{search}.


\begin{figure}[h]
\begin{codebox}
\Procname{$\proc{FFRob}({\cal D}, \Pi):$}
\li $\theta = \proc{initial-parameters}()$
\li \While \kw{True}: \Do
\li $C_*, A = \proc{sample-discretization}({\cal D}, \Pi; \theta)$
\li $\id{plan} = \proc{search}(\langle s_0, C_*, A \rangle; ...)$
\li \If $\id{plan} \neq \kw{None}$: \Then
\li \kw{return} $\id{plan}$
\End
\li $\proc{increment-parameters}(\theta)$
\End
\end{codebox}
\caption{The \ffrob{} algorithm.}\label{fig:ffrob}
\end{figure}


\section{Symbolic Planning Representation}\label{sec:rep}

We will encode robot actions that pick up and
place objects as well as move the robot in the style of traditional AI planning action
descriptions such as those shown in figure~\ref{ffrob:actions}.
In these actions, $q$, $p$, and $\gamma$ are continuous variables that range over robot
configurations, object poses, and grasp transforms, respectively.  
Because there are infinitely many values of these variables and therefore infinitely many actions,
we assume we have sampled a finite set of these values during a pre-processing phase, resulting in a finite set of actions.


\subsection{Extended Action Specification}

We model discretized \prob{} problems using a representation that extends Simple Action Specification (SAS+)~(\citealt{backstrom1995complexity}). SAS+ is expressively equivalent to \proc{strips}~(\citealt{backstrom1995complexity}) without action parameters. The key difference is that it supports variables with discrete domains instead of only propositional domains.
A generic state $s$ in SAS+ is an assignment of values to a finite set of variables ${\cal V}$. For \prob{} problems, the SAS+ variables are the same as the variables described in section~\ref{sec:formulation}. Thus, a discretized \prob{} system state is a legal SAS+ state.
For more general task and motion planning problems, the state may have additional variables such as categorical variables $v_{d_i}$ for each object that represent the cleaned or cooked status of object $o_i$ where $D_{d_i} = \{\kw{None}, \kw{Cleaned}, \kw{Cooked}\}$. 

SAS+ requires conditions and effects to be simple assignments of individual values to a subset of the variables. 

\begin{defn}
A {\em simple condition} $c \equiv [v = x]$ is a restriction that a state have value $x$ for variable $v$. 
\end{defn}

\begin{defn}
A {\em simple effect} $e \equiv v \leftarrow x$ is an assignment of a value $x$ to variable $v$. 
\end{defn}

\begin{defn}
A {\em partial state} $C = \{c_1, ..., c_k\}$ is a set of conditions.
\end{defn}

A partial state defines a set of states that satisfy its conditions. The {\em goal} of a planning problem is a partial state. 

\begin{defn}
An {\em action} $a = \langle C, E \rangle$ is a pair where $C$ is a set of simple conditions and $E$ is a set of simple effects.
\end{defn}

Even with finite domains for all the variables, there is a difficulty with determining when the robot can perform an action. In particular, $\proc{Move}$ actions have a \proc{Reachable} condition which is true if the robot can safely move from $q$ to $q'$. To concisely model and effectively plan for \prob{} problems, we need a more expressive representation that allow us to evaluate, for example, whether there exists a path between two robot configurations that does not collide with placed objects. 
We extend SAS+ by allowing conditions to be logical formulas defined on the values of the variables.
We call the resulting planning representation the {\em extended action specification} (EAS). 
This representation is also generic; however, we will focus on its application to \prob{} planning.

\begin{defn}
A {\em condition} $c \equiv f(v_{i_1}, ..., v_{i_k})$ is a restriction that a state has values for variables $v_{i_1}, ..., v_{i_k}$ that satisfy a predicate $f$.
\end{defn}

\begin{defn}
A {\em predicate} $f$ is a finite boolean combination of simple conditions.
\end{defn}

An example predicate is $f(v_{o_1}, v_{o_2}) \equiv [v_{o_1} = p_1] \vee [v_{o_2} = p_2]$, which is true when $o_1$ is currently at pose $p_1$ or $o_2$ is currently at pose $p_2$.
Let $s(v)$ give the value of variable $v$ in state $s$. 

\begin{defn}
A condition $c$ {\em holds} in a state if it evaluates to true given the values of the state's variables:
\begin{equation*}
\proc{holds}(c, s) \equiv f(s(v_{i_1}), ..., s(v_{i_k})).
\end{equation*}
\end{defn}

To concisely represent conditions sharing a common template form, we use {\em parameterized} conditions, functions from a set of parameters to a condition.
The following parameterized conditions are relevant in discretized \prob{} problems.
We use $\forall$ and $\exists$ only to compactly denote conjunctions and disjunctions over elements of our finite domains.
The parameterized condition $\proc{InReg}(o_i, R)$ has parameters composed of an object $o_i$ and a region $R \subseteq {\cal P}^{o_i}$ in its pose space. $\proc{InReg}(o_i, R)$ is true if $s(v_{o_i}) \in R$, {\it i.e.} the current placement of $o_i$ is contained within $R$. However, to express $\proc{InReg}(o_i, R)$ as a predicate, we evaluate it in the following way.
\begin{equation*}
\proc{holds}(\proc{InReg}(o_i, R), s) \equiv \exists p \in D_{o_i} \cap R. \;[s(v_{o_i}) = p].
\end{equation*}

The parameterized condition $\proc{Reachable}(q, q', (V, E))$ has parameters composed of an initial robot configuration $q$, a final robot configuration $q'$, and a discretized roadmap of robot movements $(V, E)$. $\proc{Reachable}(q, q', (V, E))$ is true if there is a collision-free path in $(V, E)$  between $q$ and $q'$, considering the positions of all fixed and movable objects as well as the object the robot might be holding and the grasp in which it is held. 
\begin{equation*}
\begin{aligned}
&\proc{holds}(\proc{Reachable}(q, q', (V, E)), s) \equiv \\
&\;\; \exists (e_1, ..., e_k) \in \proc{paths}(q, q'; (V, E)).\; \forall e \in (e_1, ..., e_k). \\ 
&\;\; \Big([s(v_h) = \kw{None}] \vee \neg \proc{robot-grasp-c}(e.\tau, s(v_h))\Big) \;\wedge \\
&\;\; \forall i \in [m]. \;\Big([s(v_{o_i}) = \kw{None}] \;\vee\\
&\;\; \big(\neg \proc{robot-obj-c}(e.\tau, (o_i, s(v_{o_i}))) \;\wedge ([s(v_h) = \kw{None}] \;\vee\\
&\;\; \neg \proc{grasp-obj-c}(e.\tau, s(v_h), (o_i, s(v_{o_i}))))\big)\Big).
\end{aligned}
\end{equation*}


Each path $(e_1, ..., e_k)$ on $(V, E)$ is composed of edges $e$ which will each have their own trajectory $e.\tau$ for moving between the incoming and outgoing vertices.
Let the predicate \proc{robot-grasp-c} be true if the robot collides with the object it is holding $v_h$, as it moves along configuration trajectory $e.\tau$.
Similarly, let \proc{robot-obj-c} be true if $o_i$ at pose $v_{o_i}$ collides with the robot along $e.\tau$, and \proc{grasp-obj-c} be true if $o_i$ at pose $v_{o_i}$ collides with the grasped object $v_h$ along $e.\tau$
We assume that roadmap $(V, E)$ is free of self-collisions or collisions with fixed obstacles as checked during its discretization. Although \proc{Reachable} is rather complicated, it still is a boolean combination of simple conditions. In section~\ref{sec:tests}, we provide a way to avoid constructing this predicate by instead directly evaluating it using an external procedure.


\begin{defn}
An action $a$ is {\em applicable} in a state $s$ if all of $a$'s conditions hold in $s$:
\begin{equation*}
\proc{applicable}(a, s) \equiv\forall c \in a.C.\; \proc{holds}(c, s).
\end{equation*}
\end{defn}

\begin{defn}
For $s, a$ such that $\proc{applicable}(a, s)$, $a$ can be {\em applied} to $s$ to produce a successor state:
\[\proc{apply}(a, s) = \begin{cases} 
v = x & (v \leftarrow x) \in a.E \\
v = s(v) & \text{otherwise}
\end{cases}\]
\end{defn}

It is often more compact to represent actions in parameterized form as {\em action schemas}. An action schema is an action with typed parameters, standing for the set of actions arising from all instantiations of the parameters over the appropriate type domains. The \proc{Pick} and \proc{Place} action schemas in figure~\ref{ffrob:actions} have parameters composed of a pose $p$, object $o_i$, grasp $\gamma$, and robot configuration $q$. The \proc{Move} action schema has parameters composed of two configurations $q, q'$ and a roadmap $(V, E)$.

\begin{figure}[h!]
\begin{codebox}
\Procname{\proc{Pick}$(p, (o_i, \gamma), q)$: }
\zi \kw{pre:} $[v_r = q]$, $[v_h = \kw{None}]$, $[v_{o_i} = p]$
\zi \kw{eff:} $v_h \leftarrow (o_i, \gamma)$, $v_{o_i} \leftarrow \kw{None}$
\end{codebox}

\begin{codebox}
\Procname{\proc{Place}$(p, (o_i, \gamma), q)$: }
\zi \kw{pre:} $[v_r = q]$, $[v_h = (o_i, \gamma)]$, $[v_{o_i} = \kw{None}]$
\zi \kw{eff:} $v_h \leftarrow \kw{None}$, $v_{o_i} \leftarrow p$
\end{codebox}

\begin{codebox}
\Procname{\proc{Move}$(q, q', (V, E))$: }
\zi \kw{pre:} $[v_r = q]$, $\proc{Reachable}(q, q', (V, E))$
\zi \kw{eff:} $v_r \leftarrow q'$
\end{codebox}
\caption{Pick, place, and move action schemas.} \label{ffrob:actions}
\end{figure}

Although we focus on \prob{} problems using these actions, we could easily define other action schemas to solve more general task and motion planning problems. For example, the \proc{Clean} and \proc{Cook} action schemas in figure~\ref{fig:additional} are useful for modeling a cooking task. The constants $R_{clean}^{o_i} \subseteq {\cal P}^{o_i}$ and $R_{cook}^{o_i} \subseteq {\cal P}^{o_i}$ are sets of poses where $o_i$ can be cleaned and cooked respectively.

\begin{figure}[h!]
\begin{codebox}
\Procname{\proc{Clean}$(o_i)$: }
\zi \kw{pre:} $[v_{d_i}= \kw{None}]$, $\proc{InReg}(o, R_{clean}^{o_i})$
\zi \kw{eff:} $[v_{d_i} \leftarrow \kw{Cleaned}]$
\end{codebox}

\begin{codebox}
\Procname{\proc{Cook}$(o_i)$: }
\zi \kw{pre:} $[v_{d_i} = \kw{Cleaned}]$, $\proc{InReg}(o, R_{cook}^{o_i})$
\zi \kw{eff:} $[v_{d_i} \leftarrow \kw{Cooked}]$
\end{codebox}

\caption{Additional clean and cook action schemas.} \label{fig:additional}
\end{figure}


\begin{defn}
An {\em EAS planning problem} $\langle s_0, C_*, A \rangle$ is specified by an initial state $s_0$, goal partial state $C_*$, and a set of actions $A$. 
\end{defn}

\begin{defn}
A finite sequence of actions $(a_1, a_2, ..., a_n) \in A \times A \times ...$ is a {\em solution} to a planning problem if and only if the corresponding sequence of states $(s_0, s_1, ..., s_n)$ starting from $s_0$ and recursively constructed using $s_i = \proc{apply}(a_i, s_{i-1})$ satisfies $\forall i \in [n]$, $\proc{applicable}(a_i, s_{i-1})$ and $\forall c \in C_*.\; \proc{holds}(c, s_n)$.
\end{defn}


\subsection{Relaxed Evaluation} \label{sec:relaxed-evaluation}

In section~\ref{sec:relaxed-planning}, it will be algorithmically advantageous to evaluate conditions in the context of {\em relaxed planning}. Relaxed planning is an approximation of standard symbolic planning which ignores delete effects~(\citealt{bonet2001planning}). Central to relaxed SAS+ planning is the notion of a relaxed state.

\begin{defn}
A {\em relaxed state} $s_+ = \{v_1 = X_1, ..., v_n = X_n\}$ is a generalized state in which each $v_i$ can simultaneously take on all values in a set $X_i \subseteq D_i$ where $X_i \neq \emptyset$.
\end{defn}

A relaxed state represents a set of states formed by all combinations of the relaxed state's values. Specifically, relaxed states can take on simultaneous values because an action in a relaxed planning problem never removes a value from the relaxed state. Thus, instead of replacing the values of variables, an action's effects add the new variable values to the relaxed state. Relaxed states are equivalent to discrete, abstracted states from Planning Modulo Theories~(\citealt{gregory2012planning}). Every state is a relaxed state; however, the converse is not true. 

A condition $c$ {\em holds} in a relaxed state $s_+$ if there exists an assignment of values in $X_i$ to each variable $v_i$ such that the condition evaluates to true. 
Let $\proc{T}(c, s_+)$ be true if such an assignment exists; then we define:

\begin{equation*}
\proc{holds}_+(c, s_+) \equiv \proc{T}(c, s_+).
\end{equation*}

Similarly, let $\proc{F}(c, s_+)$ be true if there exists an assignment of values in $X_i$ to each variable $v_i$ such that $c$ evaluates to false. Because the domain of each variable is finite, any condition can be expressed as a Boolean combination of atomic variable assignments $[v = x]$. This allows us to define $\proc{T}(c, s_+)$ by using recursion on its structure. Because it is possible for both $\proc{T}(c, s_+)$ and $\proc{F}(c, s_+)$ to hold simultaneously, we provide a recursive definition for $\proc{F}(c, s_+)$ as well. \proc{T} and \proc{F} are related in their respective recursion by negation. 

\begin{equation*}
\proc{T}(c, s_+) = \begin{cases} 
\proc{T}(c_1, s_+) \wedge \proc{T}(c_2, s_+) & c \equiv c_1 \wedge c_2 \\
\proc{T}(c_1, s_+) \vee \proc{T}(c_2, s_+) & c \equiv c_1 \vee c_2 \\
\neg \proc{F}(c', s_+) & c \equiv \neg c' \\
x_i \in s_+(v_i) & c \equiv [v_i = x_i]
\end{cases}
\end{equation*}
\begin{equation*}
\proc{F}(c, s_+) = \begin{cases} 
\proc{F}(c_1, s_+) \vee \proc{F}(c_2, s_+) & c \equiv c_1 \wedge c_2 \\
\proc{F}(c_1, s_+) \wedge \proc{F}(c_2, s_+) & c \equiv c_1 \vee c_2 \\
\neg \proc{T}(c', s_+) & c \equiv \neg c' \\
s_+(v_i) \neq \{x_i\} & c \equiv [v_i = x_i]
\end{cases}
\end{equation*}

The difference between the relaxed $\proc{holds}_+$ and standard $\proc{holds}$ is that at the atomic level, the relaxed $\proc{holds}_+$ can choose between several values of each $X_i$ to make $c$ true while the standard $\proc{holds}$ only has a single value of each $x_i$. As a consequence, when a relaxed state $s_+$ is also a standard state, $\proc{holds}_+(c, s) = \proc{holds}(c, s)$. This relaxed state condition evaluation can be seen as implementing the ``satisfies'' interface in Planning Modulo Theories~\citep{gregory2012planning} for arbitrary logical theories over discrete variables. 
 
It will be useful for the planning heuristics in section~\ref{sec:relaxed-planning} to identify a variable assignment in the relaxed state that {\em achieves} the condition.
When $\proc{holds}_+(c, s_+)$ is true, let $\proc{achievers}_+(c, s_+)$ be a similar recursive function that uses bookkeeping to identify a variable assignment that makes the condition true. 

\begin{equation*}
\proc{achievers}_+(c, s_+) = \proc{TA}(c, s_+).
\end{equation*}

Let $\proc{TA}(c, s_+)$ return a set of variable values in $s_+$ that allow $c$ to be true; and let $\proc{FA}(c, s_+)$ be the analogous set of variable values that allow $c$ to be false:

\begin{equation*}
\proc{TA}(c, s_+) = \begin{cases} 
\proc{TA}(c_1, s_+) \cup \proc{TA}(c_2, s_+) & \kern-1em c \equiv c_1 \wedge c_2 \\
\proc{any}(\proc{TA}(c_1, s_+), \proc{TA}(c_2, s_+)) & \kern-1em c \equiv c_1 \vee c_2 \\
\proc{FA}(c', s_+) & \kern-1em c \equiv \neg c' \\
\{v_i \leftarrow x_i\} & \kern-1em c \equiv [v_i = x_i]
\end{cases}
\end{equation*}
\begin{equation*}
\proc{FA}(c, s_+) = \begin{cases} 
\proc{any}(\proc{FA}(c_1, s_+), \proc{FA}(c_2, s_+)) & \kern-1em c \equiv c_1 \wedge c_2 \\
\proc{FA}(c_1, s_+) \cup \proc{FA}(c_2, s_+) & \kern-1em c \equiv c_1 \vee c_2 \\
\proc{TA}(c', s_+) & \kern-1em c \equiv \neg c' \\
\{v_i \leftarrow \proc{any}(s_+(v_i) \setminus \{x_i\})\} & \kern-1em c \equiv [v_i = x_i]
\end{cases}
\end{equation*}

The procedure $\proc{any}(x)$ arbitrarily selects an element from set $x$. Although there may be many combinations that satisfy the predicate because of \proc{any}, our algorithms just require that a single, arbitrary satisficing assignment be returned. However, the strength of these heuristics may vary depending on the assignment. \label{achievers}

\subsection{Condition Tests} \label{sec:tests}

Some conditions are computationally expensive to evaluate naively using \proc{holds}.
Consider the \proc{Reachable} condition.
Its truth value is affected by all the state variables, because the connected components of the configuration space change as the grasp and object poses change, thus affecting reachability. Holding an object changes the ``shape'' of the robot and therefore what configurations it may move between. Even more significantly, the poses of all the placed objects define additional obstacles that the robot must not collide with. For a roadmap discretization of the configuration space $(V, E)$, the \proc{Reachable} condition involves quantification over the possibly exponential number of simple paths in the discretized configuration space between $q$ and $q'$. 
Additionally, some conditions share a significant amount of logical structure. 
Consider the set of \proc{Reachable} conditions that share the same start configuration $q$.

Thus, we allow conditions to optionally be evaluated by a {\em test}, a procedure $\proc{test}(\{c_1, ..., c_n\}, s)$ which can simultaneously evaluate the predicates of conditions $\{c_1, ..., c_n\}$. 
For the \proc{Reachable} condition, we specify a test that uses dynamic programming to test if a collision-free path exists given the current state. This test evaluates all \proc{Reachable} conditions with the same start configuration $q$ at once by considering paths from $q$. 
If no \proc{test} procedure is specified, the planner defaults to the previously described methods for testing if the condition holds in standard and relaxed states.

In contrast with the semantic attachments strategy of~\cite{dornhege09icaps}, we additionally require that tests support evaluating whether relaxed states satisfy the conditions. 
This additional requirement allows the planner to include conditions evaluated by tests in heuristics to strongly guide the search. 
Once again, a test that is correct for the relaxed states is also correct for standard states, so it is sufficient to implement just $\proc{test}_+(\{c_1, ..., c_n\}; s)$. This procedure must also replace the function of $\proc{achievers}_+(c, s_+)$. Thus, in order to evaluate several conditions at once, \proc{test} returns a subset of the conditions that are true paired with their achievers. An example implementation of $\proc{test}_+$ that simply uses the default $\proc{holds}_+$ and $\proc{achievers}_+$ is the following:

\begin{equation*}
\begin{aligned}
\proc{test}_+(\{c_1, ..., &c_n\}; s_+) \equiv  \{\langle c_i, \proc{achievers}_+(s_+)\rangle \mid \\
& c_i \in \{c_1, ..., c_n\}. \;\proc{holds}(c_i; s)\}
\end{aligned}
\end{equation*}

In the case of \proc{Reachable}, the procedure \proc{test-reachable} in figure~\ref{fig:test} uses dynamic programming as well as lazy collision checking to simultaneously evaluate all \proc{Reachable} predicates from the same start configuration $q$. For each edge, a set of achievers that enable the edge to be traversed without collision is stored. Then, the full set of achievers for each end configuration $q'$ is computed by tracing back a path and taking the union of the edge achievers on the path. As a result, $\proc{test-reachable}_+$ is much more efficient than quantifying over the exponential number of paths in the roadmap $(V, E)$ for each end configuration $q'$. In practice, our \proc{test-reachable} implementation additionally memoizes the last reachable subgraph in order to avoid repeating computation for sequential evaluations in the same relaxed planning problem.




\section{Search Algorithms}\label{sec:search}

We now review existing search algorithms that can be directly applied to EAS planning problems with no modification.
The generic heuristic search procedure \proc{search} is in figure~\ref{fig:search}. The \proc{search} procedure has as arguments an EAS planning problem $\langle s_0, C_*, A \rangle$, \proc{extract} and \proc{process} procedures which alter the search control, and \proc{h} heuristic and \proc{actions} successor procedures which give a heuristic cost and generate the action successors respectively.

Depending on the behavior of the \proc{extract} and \proc{process} procedures, \proc{search} can implement many standard search control structures, including depth-first, breadth-first, uniform cost, $A^*$, greedy best-first, and hill-climbing searches. Critical to many of these strategies is a heuristic function, which maps a state $s$ in the search to an estimate of the cost to reach a goal state from $s$. We will assume that each of our actions has unit cost; however, these procedures can be easily altered when costs can be any nonnegative number. 
Appendix~\ref{search-appendix} describes the standard best-first and deferred best-first control structures used to implement \proc{extract} and \proc{process}.

\begin{figure}[h!]
\begin{codebox}
\Procname{$\proc{search}(\langle s_0, C_*, A \rangle; \proc{extract}, \proc{process}; \proc{h}, \proc{actions})$}
\li $Q = \proc{Queue}(\proc{StateN}(s_0, 0, \proc{h}(s_0), \kw{None}))$
\li \While \kw{not} \proc{empty}(\id{Q}): \Do
\li $n = \proc{extract}(Q)$
\li \If $\forall c \in C_*.\; \proc{holds}(c, n.s)$: \Do
\li \kw{return} \proc{retrace-plan}($n$)
\End
\li \For $a \in \proc{actions}(n.s, A)$: \Do
\li \If $\proc{applicable}(a, n.s)$: \Do
\li $s' = \proc{apply}(a, n.s)$
\li $\proc{process}(Q, s', n; \proc{h})$
\End\End\End
\li \kw{return} \kw{None}
\End
\end{codebox} 
\caption{The primary search control procedure.} \label{fig:search}
\end{figure}

\section{Search Heuristics}\label{sec:heuristic}

In this section, we illustrate implementations of the \proc{h} heuristic and \proc{actions} successor procedures of figure~\ref{fig:search} by adapting several existing domain-independent heuristics to EAS planning problems. 
Because they are domain-independent, this heuristics apply to any EAS planning problem, not just \prob{} problems.
However, we explore the physical interpretation of these heuristics in the context of discretized \prob{} problems. 

The simplest heuristic we consider is \proc{unsatisfied-goals}, which counts the number of unsatisfied goal conditions. 
\begin{equation*}
\proc{unsatisfied-goals}(s) = |\{c \in C_* \mid \neg \proc{holds}(c, s)\}|
\end{equation*}

Although this can be computed almost instantly, it gives an exceptionally poor estimate of the cost to the goal because the problem may require many actions to satisfy a goal condition. 
The subsequent heuristics we will discuss are more involved and consequently are able to give a much better cost estimate.

\subsection{Relaxed Planning} 
\label{sec:relaxed-planning}


Many modern domain-independent heuristics are based on solving approximate planning problems and using their solution cost as an estimate of the cost to the goal~\citep{HoffmannN01, helmert2006fast}. Although this requires more computation per search node than \proc{unsatisfied-goals}, it pays off in the search because the improved estimates significantly reduce the number of states explored. One influential planning approximation is the delete-relaxation~\cite{HoffmannN01}. As introduced in section~\ref{sec:relaxed-evaluation}, in relaxed planning problems, actions add each effect value to the state without removing the previous value.
This leads to relaxed states in which a variable can take on multiple values simultaneously. 

A solution to a relaxed planning problem is a relaxed plan, an applicable sequence of actions that results in a relaxed state that satisfies the goal conditions.
Relaxed planning problems can be solved in polynomial time, making this approximation attractive. 
However, the problem of producing a minimum length relaxed plan is NP-hard~\cite{HoffmannN01}. 

\begin{figure}[h!]
\begin{codebox}
\Procname{$\proc{compute-costs}(s, C_*, A; \proc{comb}):$} 
\li $C = C_* \cup \bigcup_{a \in A} a.C$
\li $s_+ = \{v: \emptyset \mid v \in {\cal V}\}$
\li $\id{effs}, \id{conds}, \id{acts} = \{\;\}, \{\;\}, \{\;\}$
\li $Q = \proc{Queue}(\{\proc{EffN}(v \leftarrow s(v), 0, \kw{None}) \mid \forall v \in {\cal V}\})$
\li \While \kw{not} $\proc{empty}(Q)$: \Do
\li $n = \proc{pop-min}(Q, n \mapsto n.\id{cost})$
\li \If $n.e \in \id{effs}$: \kw{continue}
\li $\id{effs}[n.e] = n$
\li $s_+(n.e.v) = s_+(n.e.v) \cup \{n.e.x\}$
\li \For $c \in C$ \If $\proc{holds}_+(c, s_+)$: \Then
\li $\id{conds}[c] = \proc{CondN}(c, \proc{achievers}(c, s_+))$ 
\li $C = C \setminus \{c\}$
\End
\li \If $C_* \subseteq \id{conds}$: \Then
\li $\id{acts}[C_*] = \proc{GoalN}(C_*, \proc{comb}(C_*; \id{effs},$ \\
\;\;\;\;\;\;\;\;\;\;\;\;\;\;\;\;\;\;\;\;\;\;\;\;\;\;\;\;\;\;\;\;\;\;\;\;\;\;\;\;\;\;\;\;\;\;\;\;\;\;\;\;\;\;\;\;\;\;\;\;\;\;\;\;\;\; $\id{conds}))$
\li \kw{return} $\id{effs}, \id{conds}, \id{acts}$
\End
\li \For $a \in A$ \If $a.C \subseteq \id{conds}$: \Do
\li $\id{acts}[a] = \proc{ActN}(a, \proc{comb}(a.C;  \id{effs},\id{conds}))$
\li $A = A \setminus \{a\}$
\li \For $e \in a.E$ \If $e \notin \id{effs}$: \Do
\li $\proc{push}(Q, \proc{EffN}(e, \id{acts}[a].\id{cost} + 1, a))$
\End\End\End
\li \kw{return} \kw{None}
\End
\end{codebox}
\caption{Method for computing relaxed planning costs.}  \label{fig:compute-costs}
\end{figure}

Despite this, several relaxed planning heuristics have been shown to give effective estimates of the distance to the goal. We will mention two of them and finally focus on the FF heuristic in particular. Each of these heuristics can be expressed using a common subroutine, \proc{compute-costs}, which is shown in figure~\ref{fig:compute-costs}. 

The delete-relaxation allows relaxed planning to be understood as a search in the space of effects rather than the space of full states. 
This is similar to a search on a hyper-graph where vertices are effects and hyper-edges are actions. At heart, \proc{compute-costs} is a version of Dijkstra's algorithm generalized for these hyper-graphs. But propagating shortest path costs in a hyper-graph differs from the traditional Dijkstra's algorithm because each hyper-edge may require reaching several vertices at a time. And there are several methods to combine the costs of reaching these vertices to produce a single cost for reaching the hyper-edge. Thus, \proc{compute-costs} requires the meta-parameter \proc{comb} which specifies the method for combining these costs. Specifically, $\proc{comb}(C; \id{eff}, \id{con})$ combines the costs of satisfying a set of conditions $C$ given the currently satisfiable effects and conditions in \id{effs} and \id{conds}.  Therefore, costs are not defined as the length of the shortest plan to reach a vertex. However, they are still a measure of the relaxed plan difficulty of reaching an effect, condition, or action. The choice of \proc{comb} will tailor \proc{compute-costs} for each heuristic.

Figure~\ref{fig:compute-costs} gives the pseudocode for \proc{compute-costs}. The inputs are a state $s$, a goal partial state $C_*$, and a set of action instances $A$. The outputs are the \id{effs}, \id{conds}, and \id{acts} maps which compose a search tree within the hyper-graph by recording cost and back pointer information. The maps are composed of effect nodes, condition nodes, and action nodes respectively. Effect nodes \proc{EffN} store an effect comprised of a variable and value, the cost at which it was first produced, and the action which achieves them. Condition nodes \proc{CondN} store a condition and a set of effects which satisfy the condition. Action nodes \proc{ActN} store an action and the cost at which the action's conditions were first satisfied.

The \proc{compute-costs} procedure starts by computing the set of unachieved conditions $C$, initializing the relaxed state $s_+$, and initializing \id{effs}, \id{conds}, and \id{acts}. It maintains a priority queue $Q$ of effects starting with each effect $v \leftarrow s(v)$ in the current state $s$. On each iteration, it pops an \proc{EffN} node $n$ from the priority queue based on its cost $n.\id{cost}$ and adds $n$ to \id{effs} and the effect $n.e = (v \leftarrow x)$ to $s_+$ if $n$ has not already been reached. Each unachieved condition $c \in C'$ then is tested to see if it can now be satisfied by the addition of $n.e$. Although not displayed in the pseudocode, this is where $\proc{test}$ procedures, which compute the truth value of several conditions at once, are also evaluated. If a condition is not satisfied, a set of effects that satisfy $c$ is returned by \proc{achievers} and recorded. If each goal condition is contained in \id{conds}, the search terminates because each goal is reachable. Here, the goal is recorded as a goal node \proc{GoalN} which is an \proc{ActN} with no effects. 
A heuristic value can then be obtained using $\id{acts}[C_*].\id{cost}$.
If the goal is not reached, we process each newly reachable and unused action $a$ by computing the cost of achieving $a$'s conditions using \proc{comb}. 
Finally, we independently push each of $a$'s unprocessed effects to the queue along with a back pointer to $a$.

This differs from existing methods for relaxed planning in SAS+ problems because conditions are allowed to be complex Boolean formulas in EAS. There may be many assignments of variables, and therefore combinations of effects, that satisfy a condition. Thus, \proc{compute-costs} explicitly searches over all the unsatisfied conditions upon reaching a new effect to determine whether any condition is now achievable. If so, it records a satisfying assignment using \proc{achievers} as described in section~\ref{achievers}. 

\begin{figure*}
\centering
\includegraphics[width=0.3\textwidth]{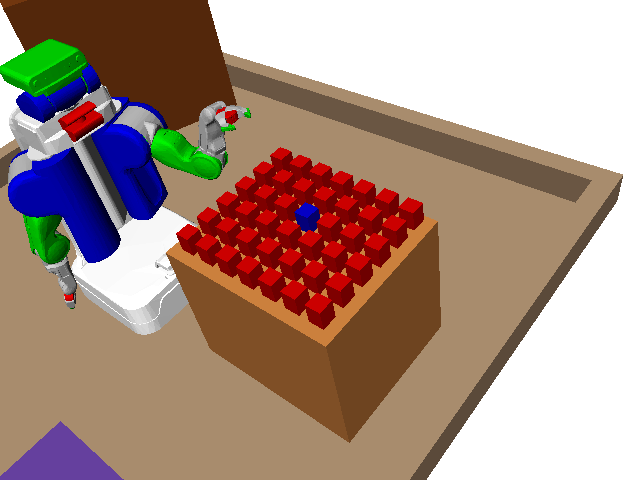}
\includegraphics[width=0.3\textwidth]{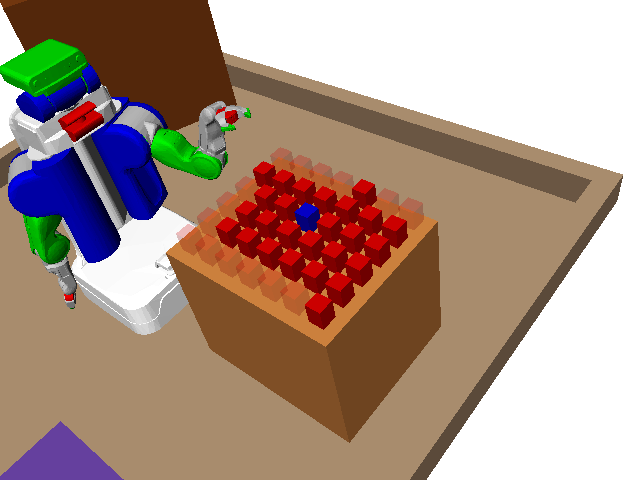}
\includegraphics[width=0.3\textwidth]{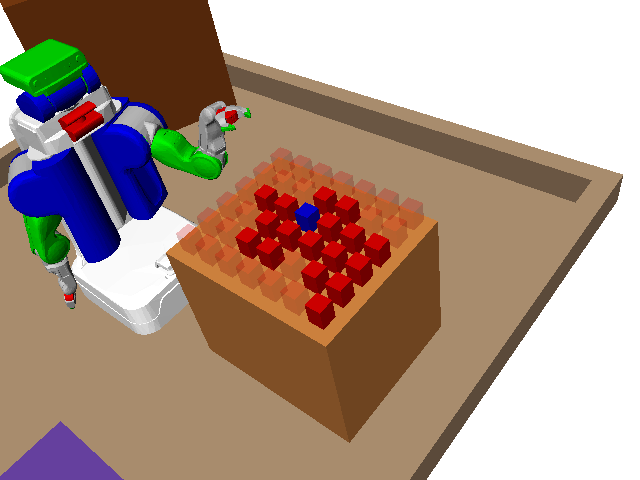}
\includegraphics[width=0.3\textwidth]{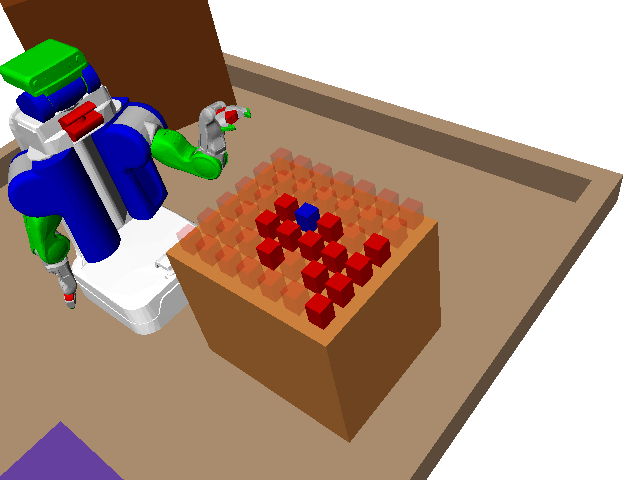}
\includegraphics[width=0.3\textwidth]{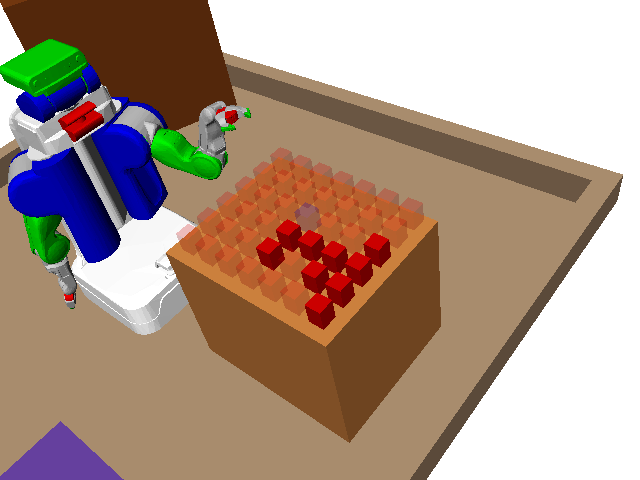}
\caption{Visualization of compute-costs for a \prob{} problem requiring picking up the blue block. Each object $o$ is made transparent at the level when $\kw{None} \in s_+(o)$. RPG levels 0, 1, and 2 are in the top row. RPG levels 3 and 4 are in the bottom row.} \label{fig:relaxed-placements}
\end{figure*}

We now describe the intuition behind \proc{compute-costs} in discretized \prob{} problems. As \proc{compute-costs} progresses, the application of each \proc{Pick}, \proc{Place}, and \proc{Move} action can be thought of creating a copy of the manipulated object or robot at the new grasp, pose, or configuration respectively. This is because the manipulated object and robot can be at many poses, grasps, and configurations simultaneously in a relaxed state. Moreover, these copies do not interfere with each other; {\it i.e.} the robot can select which of the currently available values of its configuration, its grasp, and the object poses will allow for it to perform an action. Actions that have the same cost can be viewed as being performable in parallel. Thus, the robot simultaneously tries all actions that can be performed after a relaxed plan of length $0$, then a relaxed plan of length $1$, and so on. In terms of reachability, it removes geometric constraints by removing an object from the universe when it is first picked up and never putting it back, and by assuming the hand remains empty (if it was not already) after the first \proc{Place} action. Thus, the set of satisfied $\proc{Reachable}(q, q', (V, E))$ conditions becomes increasingly larger as the procedure progresses. Figure~\ref{fig:relaxed-placements} shows the progression of objects that have greater than one pose in the relaxed state. As more objects can be picked and essentially removed from consideration, more and more reachable conditions become true.

Finally, although \proc{compute-costs} runs in polynomial time in the size of the EAS, collision checks typically dominate the runtime in \prob{} problems. 
Even if a heuristic significantly reduces the size of the main search, it might result in a net increase in computation time if it is itself too slow to compute.  Because we cache the results of collision checks, in practice, executing \proc{compute-costs} is quite fast and the heuristic functions it enables substantially reduce the number of states explored and, indeed, the total computation time.

\subsection{The HSP Heuristics}

The first two heuristics we can obtain are $h_{add}$ and $h_{max}$ which are computed by using \proc{add-comb} and \proc{max-comb} for \proc{comb} respectively. As a reminder, $C$ is a partial state and \id{eff}, \id{con} are maps of effect and condition nodes. The intention of \proc{comb} is to score the difficultly of achieving $C$ using the costs already provided in \id{eff} and \id{con}. The $h_{add}$ heuristic~\cite{bonet2001planning} returns the sum of the costs for the effects that satisfy each condition. 

\begin{equation*}
\proc{add-comb}(C; \id{eff}, \id{con}) = \sum_{c \in C}\sum_{e \in \id{con}[c].E} \id{eff}[e].\id{cost}
\end{equation*}

This heuristic is optimistic, in the sense that if the delete effects were taken into account, it might take more steps to achieve each individual goal from the starting state; it is also pessimistic, in the sense that the actions necessary to achieve multiple goal conditions might be ``shared.'' 

An admissible heuristic, $h_{max}$~\cite{bonet2001planning}, is obtained by taking the maximum of the costs of the goal literals, rather than the sum. But $h_{max}$ is found in practice to offer weaker guidance.

\begin{equation*}
\proc{max-comb}(C; \id{eff}, \id{con}) = \max_{c \in C}\max_{e \in \id{con}[c].E} \id{eff}[e].\id{cost}
\end{equation*}

%
%
%

\subsection{The FF Heuristic} \label{backchain}\label{heuristic} \label{rpg} 

The FF heuristic $h_{\it ff}$~\cite{HoffmannN01} derives its heuristic cost from the length of a relaxed plan. 
The relaxed plan is computed by first calculating either $h_{max}$ or $h_{add}$ using \proc{compute-costs} to produce an ordering of effects and actions as represented by $\id{effs}, \id{conds}, \id{acts}$. 
Then, \proc{extract-relaxed-plan} performs a backwards pass to identify a relaxed plan from $\id{effs}, \id{conds}, \id{acts}$.

In the event where $h_{max}$ is used to produce the ordering, $\id{effs}, \id{conds}, \id{acts}$ represent a {\em relaxed plan graph} (RPG).
An RPG is a sequence of alternating {\em layers} of effects and actions. The first layer consists of all effects that are true in the starting state.  Action layer $i$ contains all actions whose conditions are present and simultaneously achievable in effect layer $i$. Effect layer $i+1$ contains all effects that are possibly achievable after $i$ actions. The depth of an effect or action in the RPG is equal to its $h_{max}$ cost (assuming unit action costs). 
In the event where $h_{add}$ is used to produce the ordering, the resulting structure is not semantically an RPG; however, the resulting structure functions akin to an RPG and can even lead to tighter heuristic estimates.

FF performs an efficient backward-chaining pass in the RPG to determine which actions, if they could be performed in parallel without delete effects, would be necessary to achieve the goal conditions. An important advantage of the FF heuristic is that it does not over-count actions if one action achieves multiple effects. 

\begin{figure}[h]
 
\begin{equation*}
\proc{easiest}(C, \id{goals}; \id{eff}, \id{con}) = \sum_{c \in C}\sum_{\begin{subarray}{l}e \in \id{con}[c].E \\e \notin \id{goals}\end{subarray}} \id{eff}[e].\id{cost}
\end{equation*}


\begin{codebox}
\Procname{$\proc{extract-relaxed-plan}(s, C_*, \id{effs}, \id{conds}, \id{acts}):$} 
\li $n = \id{acts}[\langle C_*, \emptyset \rangle].\id{cost}$
\li $\id{goals} = \{i: \emptyset \mid \forall i \in [n]\}$
\li \For $c \in C_*$ \For $e \in \id{conds}[c].E$: \Do 
\li $\id{goals}[\id{effs}[e].\id{cost}] \;\cup\!= \{e\}$
\End
\li $\id{plan} = \emptyset$
\li \For $i \in (n, ...,1)$: \Do
\li \For $e \in \id{goals}[i]:$ \Do 
\li $A_e = \{a \in \id{acts} \mid e \in a.E$ \\
\;\;\;\;\;\;\;\;\;\;\;\;\;\;\;\;\;\;\;\;\;\;\;$\id{acts}[a].\id{cost} = \id{effs}[e].\id{cost}-1\}$
\li $a = \arg\!\min_{a \in A_e}\proc{easiest}(a.C, \id{goals}; \id{eff},$ \\ 
\;\;\;\;\;\;\;\;\;\;\;\;\;\;\;\;\;\;\;\;\;\;\;\;\;\;\;\;\;\;\;\;\;\;\;\;\;\;\;\;\;\;\;\;\;\;\;\;\;\;\;\;\;\;\;\;\;\;\;\;\;\;\;\;\;\;\;\;\;$\id{conds})$ 
\li $\id{plan} \;\cup\!= \{a^*\}$
\li \For $c \in a^*.C$ \For $e' \in \id{conds}[c].E$: \Do
\li $\id{goals}[\id{effs}[e'].\id{cost}] \;\cup\!= \{e'\}$
\End
\li \For $e' \in a^*.E$: \Do
\li $\id{goals}[\id{effs}[e'].\id{cost}] \;\setminus\!= \{e'\}$
\End\End\End
\li \kw{return} $\id{plan}$
\End
\end{codebox}
\caption{Method for extracting a relaxed plan.}\label{fig:extract-plan}
\end{figure}

The \proc{extract-relaxed-plan} procedure in figure~\ref{fig:extract-plan} extracts a relaxed plan from the RPG. 
The plan extraction procedure greedily processes each layer backwards from $i = (n, ..., 1)$, starting with the set of effects that achieve the goal conditions $C_*$. For each effect $e \in \id{goals}[i]$ identified as a goal on the $i$th layer of the RPG, it seeks the ``cheapest'' action $a^*$ that can achieve it using \proc{easiest}. The minimizing $a^*$ is added to the relaxed plan $\id{plan}$, $e'$ and any other effects achieved by $a$ on the current layer are removed from \id{goals}, and the conditions $a^*.C$ are processed and their satisfying effects are added to \id{goals}.  This process continues until each layer is processed. Once finished, the FF heuristic returns the number of actions in $\id{plan}$.

Our formulation of \proc{extract-relaxed-plan} is very similar to the original FF \proc{extract-relaxed-plan} algorithm. 
The modification to support EAS planning problems is simply performed, given that the effect achievers have been computed for each condition, by replacing conditions with the effects that satisfy them. 
Our metric for choosing the cheapest action is different from the original formulation, though. 
The original easiest-action metric uses the $h_{add}$ cost of each action which is separately computed before \proc{extract-relaxed-plan}. 
Our \proc{easiest} procedure uses the original cost and additionally discounts the cost of goals that have already been identified by addition to \id{goals}. 
The intuition for this change is that actions that do not add many new goals are greedily good choices.

As our results in section~\ref{sec:exp} show, $h_{\it ff}$ has the best performance in our \prob{} experiments. 
Our intuition behind why $h_{\it ff}$ performs well for \prob{} problems comes from its backwards pass. 
There are often several ways to achieve a \prob{} goal. For example, consider the set of grasps and approach trajectories to pick up an object. Many of these approaches involve paths that require moving a different set of objects. And many of these paths may be performable on the same layer of the RPG. This is typically the case when an object is surrounded by an approximately even distribution of objects on several of its sides. The backwards pass, particularly through its greedy discounting, will frequently select the approach that will require moving the fewest of additional objects given the choices it has already made. Thus, the resulting relaxed plan usually reuses goals and gives a better estimate of the optimal cost to the goal. 


\subsection{Helpful Actions}

We now describe the implementation of the \proc{actions} procedure used by the heuristic search algorithm in figure~\ref{fig:search}. The simplest implementation is just to return the full set of actions $A$. However, we allow the optional specification of \proc{actions} to be a function that can more efficiently return the set of applicable actions. For \prob{} problems, we compute the set of reachable configurations all at the same time using a procedure similar to \proc{test-reachable} in order to determine the applicable move actions.


Additionally, because $h_{\it ff}$ computes a relaxed plan, we can use the resulting plans to prune and order a set of {\em helpful actions} for a given state. 
Helpful action pruning strategies reduce the choice of the next action to those that were identified to be achieve goals on the relaxed plan via the heuristic computation. 
Additionally, they can order the application of these actions such that actions deemed more helpful are attempted first. Both of these methods can reduce the effective branching factor and the number of future heuristic computations needed in the search. But these pruning methods sacrifice completeness at the expense of strongly improved search performance. Completeness can be recovered though by using multiple priority queues as introduced by~\cite{helmert2006fast}. We use a version of the helpful actions strategy that returns the {\em first actions} followed by the {\em first goal achievers}. The first actions are actions on a relaxed plan that are immediately performable. The first goal achievers are actions that have an effect on the first layer of a relaxed plan. Thus, the first actions are a subset of the first goal achievers. 

%


\section{Discretization}\label{sec:rob}

In order to use the EAS representation and planners, we still have to discretize a \prob{} problem in a pre-processing phase by sampling a finite number of actions from each action schema. 

\begin{defn}
A {\em discretized \prob{} domain} $D = \langle (V, E), \{(P^{o_1}, G^{o_1})..., (P^{o_m}, G^{o_m})\} \rangle$ is specified by a robot configuration roadmap $(V \subseteq {\cal Q}, E \subseteq V \times V)$ as well as a finite sets of placements $P^{o_i} \subseteq {\cal P}^{o_i}$ and grasps $G^{o_1} \subseteq {\cal G}^{o_i}$ for each of the $m$ moveable objects $o_i$. 
\end{defn}

In figure~\ref{fig:discretization}, the procedure \proc{sample-discretization} both samples a discretized \prob{} problem and produces an EAS specification of the goal and actions for the discretized problem. First, the goal set of states $S_*$ is converted into a set of predicates $C_*$ using \proc{convert-goal}. 
Sets of $\proc{Pick}$ and $\proc{Place}$ actions are sampled using $\proc{s-pick-place}$ according to the current parameter vector $\theta$. A set of $\proc{Move}$ actions is constructed using $\proc{s-move}$ by creating a roadmap that connects the start configuration $q_0$, sampled goal configurations, $Q_*$, and $\proc{Pick}$ and $\proc{Place}$ configurations. Finally, \proc{sample-discretization} returns $C_*$ as well as the combined set of actions $A$.

\begin{figure}[h]
\begin{codebox}
\Procname{$\proc{sample-discretization}({\cal D}, \Pi; \theta):$}
\li $C_* = \proc{convert-goal}(S_*)$ 
\li $A_{pick}, A_{place}$= $\proc{s-pick-place}({\cal D}, \Pi; \theta)$
\li $A _{move} = \proc{s-move}({\cal Q}, q_0, Q_*, A_{pick}, A_{place}; \theta)$
\li $A = A_{pick} \cup A_{place} \cup A_{move}$
\li \kw{return} $C_*, A$
\end{codebox}
\caption{Procedure for sampling a discretized \prob{} problem and producing an EAS problem specification.}\label{fig:discretization}
\end{figure}

The rest of this section gives the details of \proc{s-pick-place} and \proc{s-move} as well as the procedure to evaluate complex \proc{Reachable} predicates.
We start by sampling the \proc{Pick} and \proc{Place} action schemas.
Recall that \proc{Pick} and \proc{Place} have a condition that the robot be at a specific configuration in order to execute the pick or place. These configurations will serve as target configurations when sampling the roadmap $(V, E)$. \proc{Move} actions are created from pairs of these target configurations.

Sampling \proc{Pick} and \proc{Place} actions has a deep connection to sampling the modes of the system~(\citealt{Hauser, HauserIJRR11}). Specifically, the collection of poses and grasps will define the set of transit and transfer modes reachable from $s_0$. The \proc{Move} actions then represent discretized motion plans within a mode. We will later revisit this idea in the theoretical analysis of \ffrob{} in section~\ref{sec:theory}.

The number of samples chosen for each sampling procedure depends on a parameter vector $\theta$. 
As previously said, $\theta$ will be iteratively increased in the event that not enough samples were chosen to find a solution.
Additionally, to test whether this sampled set of poses, grasps, and configurations could possibly contain a plan, we can compute the heuristic value of the starting state $s_0$ using the actions derived from the samples, as described in section~\ref{heuristic}. 
If it is infinite, meaning that the goal is unreachable even under extremely optimistic assumptions, then we return to this procedure and draw a new set of samples. Note that although a finite heuristic value is necessary for a plan to exist, it is not sufficient. Thus, the search algorithm may report that the set of samples is not sufficient to solve the problem although $s_0$ had a finite heuristic cost. 
However, in our experiments, this finite heuristic test can save a significant amount of time because computing a heuristic cost is much less expensive than solving the EAS planning problem.

Finally, evaluating \proc{Reachable} still requires performing many expensive collision checks in the context of many different robot grasps and poses of the objects.  We address this problem by using a shared {\em roadmap} data structure called a {\em conditional reachability graph} (\crg{}). The \crg{} is graph $(V, E)$, which is related to a PRM~(\citealt{Kavraki96}), that answers reachability queries, conditioned on the poses of objects and the robot's grasp, by lazily computing answers on demand and caching results to speed up future queries.

\subsection{Pick and Place Actions}

\begin{figure}
\includegraphics[width=0.49\textwidth]{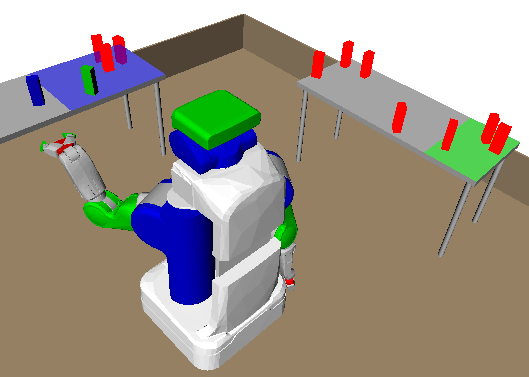}
\caption{Sampled placements for the blue and green blocks.} \label{fig:placements}
\end{figure}

Figure~\ref{fig:transfers} gives the procedure for sampling action instances from the \proc{Pick} and \proc{Place} action schemas. This procedure, for each object, first produces poses and grasps useful for the problem, by using initial poses and grasps and sampling goal poses and grasps (if available) as well as uniformly sampling each space. For our implementation of $\proc{s-poses}$ and $\proc{s-grasps}$, we sample values using uniform rejection sampling (rejecting, for example, poses that collide with fixed obstacles); however, this can be done in other ways such as by choosing evenly spaced samples. Figure~\ref{fig:placements} shows sampled placements in red for the blue and green blocks. Additional placements are sampled for the blue and green goal regions.

\begin{figure}
\includegraphics[width=0.49\textwidth]{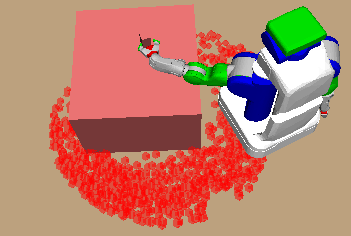}
\caption{The inverse reachability database transformed relative to the grasp transform.} \label{fig:ir}
\end{figure}

Once the poses and grasps ($p$ and $\gamma$) are sampled, \proc{s-pick-place} uses \proc{s-ik} to sample valid configurations of the robot base and manipulator that reach the end-effector pose of $p \times \gamma^{-1}$. We implement  \proc{s-ik} by first sampling base poses from a precomputed inverse reachability database that are nearby the end-effector pose as shown in figure~\ref{fig:ir}. Then, we sample collision-free, analytical inverse kinematics (IK) solutions using ikfast~\cite{diankov2010automated}. Finally, each pose, grasp, and configuration tuple are used as the arguments of a $\proc{Pick}$ and $\proc{Place}$ action instance. These actions are added the respective sets of EAS actions, $A_{pick}$ and $A_{place}$.

\begin{figure}[h]
\begin{codebox}
\Procname{$\proc{s-pick-place}({\cal D}, \Pi; \theta):$}
\li $A_{pick}, A_{place} = \emptyset, \emptyset$
\li \For $i \in [m]$: \Do
\li $P^{o_i} = \{p_0^{o_i}\} \cup \proc{s-poses}(P_*^{o_i}; \theta) \cup \proc{s-poses}({\cal P}^{o_i}; \theta)$
\li $G^{o_i} = \proc{s-grasps}({\cal G}^{o_i}; \theta)$
\li \If $i = h_0: G_o \;\cup\!= \{g_0^{o_{h_0}}\}$
\li \If $i = h_*: G_o \;\cup\!=  \proc{s-grasps}(G_*^{o_{h_*}}; \theta)$
\li \For $(p, g) \in P^{o_i} \times G^{o_i}$: \Do
\li $\gamma = \proc{transform}(g)$
\li \For $q \in \proc{s-ik}({\cal Q}, p \times \gamma^{-1}; \theta)$: \Do
\li $A_{pick} \;\cup\!= \{\proc{Pick}(p, (o_i, \gamma), q)\}$
\li $A_{place} \;\cup\!= \{\proc{Place}(p, (o_i, \gamma), q)\}$
\End\End\End
\li \kw{return} $A_{pick}, A_{place}$
\end{codebox}
\caption{Procedure for sampling the $\proc{Pick}$ and $\proc{Place}$ actions.}\label{fig:transfers}
\end{figure}

\subsection{Conditional Reachability Graph}

Using the \proc{Pick} and \proc{Place} robot configurations as targets, 
we now sample robot movements and trajectories that will reach these pick and place configurations. 
In the process, we will create a {\em conditional reachability graph} (\crg{}).
The \crg{} is a partial representation of the connectivity of the space of sampled
configurations, conditioned on the placements of movable objects as
well as on what is in the robot's hand. It efficiently allows us to evaluate $\proc{Reachable}(q1, q2, (V, E))$ predicates.
It is similar in spirit to the roadmaps of~\cite{Leven} in that it is
designed to support solving multiple motion-planning queries in
closely related environments. Formally, it is a graph $(V, E)$ where
the {\em vertices} $V$ are a set of robot configurations, $q \in V$. 
The {\em edges} $E$ are triplets $e = \langle q, q', \tau \rangle$ where $q, q'$ are pairs of vertices and $\tau$ is a trajectory that connects $q$ and $q'$. 
Each edge is also annotated with an initially empty map of {\em validation} conditions of the form $e.\id{valid}[\langle \rho, g \rangle] = b$ where $b = \kw{True}$ if the edge is traversable for a placement of object $\rho$ and grasp $g$ when there are no other objects in the world. 
Otherwise, $b = \kw{False}$. There are three cases for a $\langle \rho, g \rangle$ pair:  
\begin{itemize}
\item $\langle (o, p), \kw{None} \rangle$: safe for the robot to traverse $e$ when object $o$ is placed at pose $p$ and the robot is not holding any object.
\item $\langle \kw{None}, (o', \gamma) \rangle$: safe for the robot to traverse $e$ when no object is placed and the robot is holding object $o'$ with grasp transform $\gamma$.
\item $\langle (o, p), (o', \gamma) \rangle$: safe for the robot to traverse $e$ when object $o$ is placed at pose $p$ and the robot is holding object $o'$ with grasp transform $\gamma$.
\end{itemize}

\begin{figure}
\includegraphics[width=0.49\textwidth]{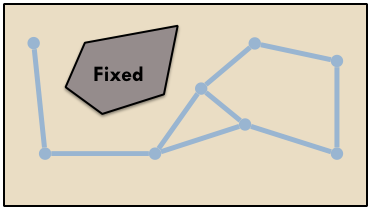}
\includegraphics[width=0.49\textwidth]{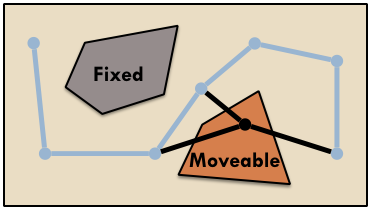}
\caption{An unconditioned \crg{} and the same \crg{} conditioned on the placement of a single moveable object.} \label{fig:crg}
\end{figure}

The validation conditions on the edges are not pre-computed; they will be computed lazily, on demand, and cached in this data structure. These conditions are separated in this way in order to maximize the amount of collision caching used. Note that the procedure for determining $e.\id{valid}[\langle (o, p), (o', \gamma) \rangle]$ does not need to compute whether the robot collides with either $o$ or $o'$ because those conditions will already have been computed and stored in $\langle (o, p), \kw{None} \rangle$ and $\langle \kw{None}, (o', \gamma) \rangle$ respectively. However, it still will need need to compute whether $(o, p)$ collides with $(o', \gamma)$. Figure~\ref{fig:crg} depicts a cartoon \crg{}. The bottom figure is conditioned on a moveable object which temporarily removes three edges from the traversable roadmap.

\subsection{Constructing the \crg{}}

The \crg{} is initialized in the pre-processing phase when sampling the $\proc{Move}$ actions. The \proc{s-move} procedure is outlined in figure~\ref{crg}. The initial configuration and goal configurations sampled using \proc{s-configs} are used to initialize the \crg{}. It then calls the \proc{s-approaches} procedure which generates trajectories for moving near each \proc{Pick} and \proc{Place} configuration. 

Let $\proc{parameters}(a)$ give the tuple of parameters for action instance $a$. For each \proc{Pick} and \proc{Place} configuration, \proc{s-approaches} samples a nearby configuration $q'$ using $\proc{s-nearby-config}$ for the purpose of approaching the \proc{Pick} and \proc{Place} configuration $q$.
In our implementation, we do this concatenating the previous base pose with a predetermined manipulator configuration used when carrying objects. 
Then it calls \proc{s-appr-traj}, which samples trajectories $\tau$ between $q$ and $q'$. For our implementation, we call RRT-Connect ~(\citealt{KuffnerLaValle}) between $q$ and $q'$ in the configuration space of just the manipulator (because we use the same base pose). We disallow trajectories that either collide with fixed obstacles while holding $o$ at grasp $g$ or collide with $o$ at pose $p$ when moving with an empty hand in order to use the trajectory for both \proc{Pick} and \proc{Place} actions. Our implementation assumes the robot is holonomic, so trajectories are reversible. For non-holonomic robots, a separate trajectory from $q'$ and $q$ must be sampled. An edge following the trajectory and an edge following the reversed trajectory are then added to the \crg{}. 

\begin{figure}
\includegraphics[width=0.49\textwidth]{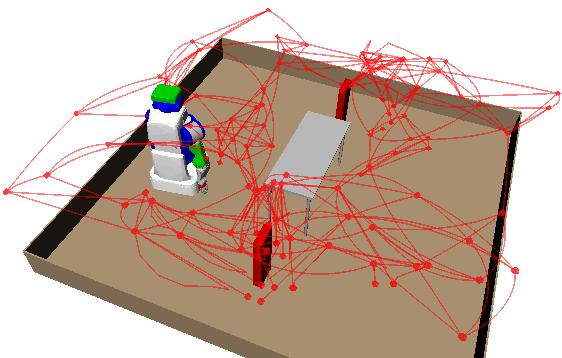}
\caption{Full \crg{} visualized using end-effector poses.} \label{fig:prm-crg}
\end{figure}

Next, \proc{s-move} calls the \proc{s-roadmap} procedure which attempts to connect the configurations in the \crg{}. The procedure described in figure~\ref{crg} is for a fixed-degree PRM. It samples additional roadmap configurations using \proc{s-configs} to attempt to connect the roadmap. For each configuration in the roadmap, \proc{s-roadmap} attempts to connect the configuration to its nearest neighbors in $V$ given by \proc{nearest-neighbors}. It uses \proc{s-traj} to linearly interpolate between configurations. The number of additional configurations to sample and the desired degree of the PRM are given by the parameter vector $\theta$. Figure~\ref{fig:prm-crg} shows a sampled \crg{} for a NAMO problem. 

\begin{figure}
\includegraphics[width=0.49\textwidth]{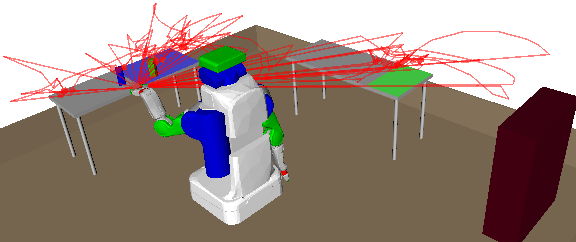}
\caption{A star-graph \crg{} visualized using end-effector poses.} \label{fig:star-crg}
\end{figure}

In cases where domain-dependent information is available, it may make sense to use a different \proc{s-move} procedure. For example, for problems where objects can only be placed on tables, it is wasteful to create a dense roadmap for moving between tables because placed objects cannot possibly affect the validity of edges at a certain distance away from the tables. In our experiments involving objects that can only be placed on tables, we use a version of \proc{s-roadmap} that is sparse away from tables while still dense nearby tables by creating a star-graph of trajectories that connect an arbitrary reference configuration (such as $q_0$) to configurations near each table as shown in figure~\ref{fig:star-crg}.

Finally, \proc{s-move} creates a \proc{Move} action instance for pairs of configurations $V$ that are connected in the roadmap as tested by a standard breadth-first search using $\proc{path}(q, q'; (V, E))$ (without considering any placed or held objects). In practice, we only create \proc{Move} actions between start and goal configurations as well as configurations at which the robot can perform a \proc{Pick} or \proc{Place} because the robot will never need to stop at an intermediate configuration.
Lastly, \proc{s-move} returns the set of move actions $A_{move}$.


\begin{figure}[h]
\begin{codebox}
\Procname{$\proc{s-approaches}({\cal Q}, (V, E), A_{pick}, A_{place}; \theta):$}
\li \For $a \in (A_{pick} \cup A_{place})$: \Do 
\li $(p, g, q) = \proc{parameters}(a)$
\li \For $q' \in  \proc{s-nearby-config}({\cal Q}, q; \proc{sch}(a), p, g; \theta)$ \Do
\li $V \;\cup\!= \{q, q'\}$
\li \For $\tau \in \proc{s-appr-traj}({\cal Q}, q, q'; \proc{sch}(a), p, g; \theta)$ \Do
\li $E \;\cup\!= \{(q, q', \tau), (q', q, \proc{reverse}(\tau))\}$
\End\End\End
\end{codebox}

\begin{codebox}
\Procname{$\proc{s-roadmap}({\cal Q}, (V, E); \theta):$}
\li $V \;\cup\!= \proc{s-configs}({\cal Q}; \theta)$
\li \For $q \in V$: \Do
\li \For $q' \in \proc{nearest-neighbors}(V, q; \theta)$: \Do
\li \For $\tau \in \proc{s-traj}({\cal Q}, q, q'; \theta)$ \Do
\li $E \;\cup\!= \{(q, q', \tau), (q', q, \proc{reverse}(\tau))\}$
\End\End\End
\end{codebox}

\begin{codebox}
\Procname{$\proc{s-move}({\cal Q}, q_0, Q_*, A_{pick}, A_{place}; \theta):$}
\li $(V, E) = (\{q_0\} \cup \proc{s-configs}(Q_*; \theta), \emptyset)$
\li $\proc{s-approaches}({\cal Q}, (V, E), A_{pick}, A_{place}; \theta)$
\li $\proc{s-roadmap}({\cal Q}, (V, E); \theta)$
\li $A_{move} = \{\proc{Move}(q, q', (V, E)) \mid (q, q') \in V \times V,$ \\
$\;\;\;\;\;\;\;\;\;\;\;\;\;\;\;\;\proc{path}(q, q'; (V, E)) \neq \kw{None}\}$
\li \kw{return} $A_{move}$
\end{codebox}

\caption{Procedures for constructing the \crg{}.} \label{crg}
\end{figure}

\subsection{Querying the \crg{}} \label{test}

\begin{figure}
\includegraphics[width=0.49\textwidth]{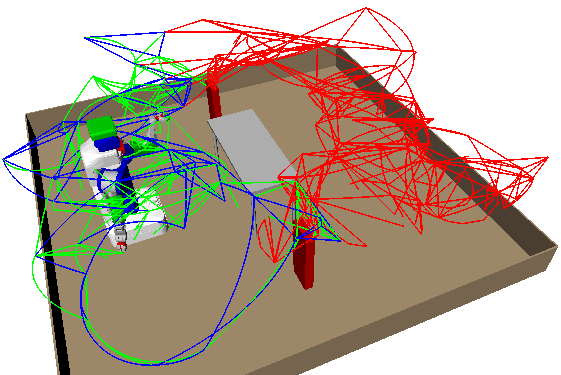}
\caption{The reachable \crg{}. Green edges are on the BFS tree. Blue edges have reachable vertices but were not selected for the BFS tree. Red edges have at least one unreachable vertex.} \label{fig:reachable-crg}
\end{figure}

Now that we have a \crg{}, we can use it to test whether \proc{Reachable} conditions hold in relaxed state $s_+$, as shown in $\proc{test-reachable}$ in figure~\ref{fig:test}. Recall that in each relaxed state $s_+$, each object can simultaneously be: missing entirely, in multiple poses, and in multiple grasps.
Similarly, the hand can hold several objects while also remaining empty.
We need to determine whether there is some simultaneous assignment of all these variables using the values present in $s_+$ that allows a legal path from a start $q$ to $q' \in V$. The test constructs a subgraph $(V, E')$ of the \crg{} that consists only of the edges that are each independently valid for some choice of object poses and robot grasps from $s_+$.
Additionally, each edge $e$ is augmented with a temporary set of these pose and grasp achieving values $e.{\id ach}$ that collectively allow collision-free traversal of the edge. The test then searches that graph to see if configuration $q'$ is reachable from $q$. It calls two sub procedures: \proc{test-grasp} and \proc{test-obj}. Figure~\ref{fig:reachable-crg} displays the set of reachable \crg{} configurations from the current robot configuration given the placements of the movable objects for a NAMO problem.

The procedure \proc{test-grasp} checks, given an edge $e$, the relaxed state $s_+$, and a placement $\rho$ (which may be \kw{None}), whether the edge is valid for some grasp $g$ in the relaxed state. If $\kw{None}$ is a grasp in $s_+$, the \proc{test-grasp} can immediately report success because it can navigate the edge without colliding with the fixed obstacles (assuming the robot itself does not collide with $\rho$). Otherwise, it processes each grasp in $s_+$. If $\langle \rho, g \rangle$ has not already been cached as a validation condition, it is computed. If $\rho = \kw{None}$, the procedure computes whether the robot moving along $e.\tau$ with grasp $g$ either causes a self-collision or $g$ collides with fixed obstacles using $\proc{robot-grasp-c}$. Otherwise when $\rho = (o, p)$, the procedure computes whether the grasp $g$ moving along $e.\tau$ causes a collision with object $o$ at pose $p$ using $\proc{grasp-obj-c}$. Then, it can check $e.\id{valid}$ to obtain the cached answer to the query. If \proc{test-grasp} finds a satisfying grasp, it returns \kw{True}. Otherwise, it returns \kw{False}.

Similarly, the procedure \proc{test-obj} computes, for an edge $e$, the relaxed state $s_+$, and an object $o$, whether the edge is valid for some pose $p$ in the relaxed state. If $\kw{None}$ is a pose for object $o$ in $s_+$, the procedure can immediately report success because it can navigate the edge when $o$ is not placed. Otherwise, it processes each pose for $o$ in $s_+$. If $\langle (o, p), g \rangle$ has not already been cached as a validation condition, it is computed. The procedure computes whether the robot moving along $e.\tau$ causes a collision with $o$ at pose $p$ using $\proc{robot-obj-c}$. It then checks $e.\id{valid}$ to obtain the cached answer to the query and calls \proc{test-grasp} with placement $\rho = (o, p)$ to see if this pose admits a legal grasp. If \proc{test-obj} finds a satisfying pose, it returns \kw{True}. Otherwise, it returns \kw{False}.


Many of these checks can be done very efficiently using simple bounding-box computations. Additionally, the \crg{} caches the polyhedral structure of the robot along the trajectory, which can speed up later collision checks along the same edge by not having to place the robot a second time. Within \proc{test-reachable}, if there is not a valid grasp and pose for each object that allows safe passage across an edge, the edge is removed from the subgraph $E'$. Finally, $\proc{bfs-tree}(q; (V, E'))$ performs a standard breadth-first search through the subgraph to construct a BFS tree $(N, T)$. \proc{test-reachable} the returns the subset of $\{c_1, ..., c_n\}$ which have end configuration $q'$ in $(N, T)$. Additionally, a set of achiever variable values is computed using $\proc{trace-ach}(c_i.q'; (N, T))$ by tracing a path to $q$ and taking the union of $e.{\id ach}$ for visited edges.

Recall that for standard non-relaxed states, each object can only be at one pose at a time, and the robot can hold at most one object, so, the loops over poses $s_+(v_{o_i})$ and grasps $s_+(v_h)$ will only process a single pose or grasp respectively. This test is also efficient for relaxed states that arise during the relaxed planning process. As soon as an object $o_i$ is picked up, it will have $\kw{None} \in s_+(v_{o_i})$, and no further collision checks using object $o_i$ will be required. Additionally, before a new object pose or grasp can be added to the relaxed state, a $\kw{None}$ pose or grasp must be added from a \proc{Pick} or \proc{Place} action respectively. Thus, $s_+(v_{o_i})$ and $s_+(v_h)$ will either only contain a single pose or grasp or they will contain $\kw{None}$, which expedites the test.

Finally, our implementation of \proc{test-reachable} actually tests the validity of the graph lazily while performing the search. We also exploit the fact that we are only interested in the connected component of the \crg{} that includes the current robot configuration which further increases efficiency. Moreover, for sequential evaluations for the same heuristic, \proc{test-reachable} expands on the previously reachable subgraph to avoid reevaluating traversable edges.

\begin{figure}[h!]

\begin{codebox}
\Procname{$\proc{test-grasp}(s_+, e, \rho):$}
\li \If $\kw{None} \in s_+(v_h)$: \Then 
\li $e.{\id ach} \;\cup\!= \{v_h \leftarrow \kw{None}\}$
\li \kw{return} \kw{True}
\End
\li \For $g \in s_+(v_h)$: \Do
\li \If $\langle \rho, g \rangle \notin e.\id{valid}$: \Then
\li $e.\id{valid}[\langle \rho, g \rangle]$ = $(\rho = \kw{None}$ \kw{and} \\
\;\;\;\;\;\;\;\;\;\;\;\;\;\;\;\;\kw{not} $\proc{robot-grasp-c}(e.\tau, g))$ \kw{or} \\
\;\;\;\;\;\;\;\;\;\;\;\;\;\;\;\;$(\rho \neq \kw{None}$ \kw{and} \kw{not} $\proc{grasp-obj-c}(e.\tau, g, \rho))$
\End
\li \If $e.\id{valid}[\langle \rho, g \rangle]$: \Then 
\li $e.{\id ach} \;\cup\!= \{v_h \leftarrow g\}$
\li \kw{return} \kw{True}
\End\End
\li \kw{return} \kw{False}
\end{codebox}

\begin{codebox}
\Procname{$\proc{test-obj}(s_+, e, o):$}
\li \If $\kw{None} \in s_+(v_o)$: \Then
\li $e.{\id ach} \;\cup\!= \{o \leftarrow \kw{None}\}$
\li \kw{return} \kw{True}
\End
\li \For $p \in s_+(v_o)$: \Do
\li \If $\langle (o, p), \kw{None} \rangle \notin e.\id{valid}$: \Then
\li $e.\id{valid}[\langle (o, p), \kw{None} \rangle]$ = \\ 
\;\;\;\;\;\;\;\;\;\;\;\;\;\;\;\;\kw{not} $\proc{robot-obj-c}(e.\tau, (o, p))$
\End
\li \If $e.\id{valid}[\langle (o, p), g \rangle]$ \kw{and} $\proc{test-grasp}(s_+, e, (o, p))$: \Then
\li $e.{\id ach} \;\cup\!= \{o \leftarrow p\}$
\li \kw{return} \kw{True}
\End\End\End
\li \kw{return} \kw{False}
\end{codebox}


\begin{codebox}
\Procname{$\proc{test-reachable}(\{c_1, ..., c_n\}, s_+; (V, E)):$}
\li $q = c_1.q$
\li $E' = E$
\li \For $e \in E$: \Do
\li \If \kw{not} $\proc{test-grasp}(s_+, e, \kw{None})$: \Then
\li $E' = E' \setminus \{e\}$
\li \kw{continue}
\End
\li \For $i \in [m]$: \Do
\li \If \kw{not} $\proc{test-obj}(s_+, e, o_i)$: \Then 
\li $E' = E' \setminus \{e\}$
\li \kw{break} 
\End\End\End
\li $(N, T) = \proc{bfs-tree}(q; (V, E'))$
\li \kw{return} $\{\langle c_i, \proc{trace-ach}(c_i.q'; (N, T))  \rangle \mid c_i.q' \in N\}$
\end{codebox}
\caption{Procedure for querying the \crg{}.}  \label{fig:test}
\end{figure}


\section{Review of sPRM Theoretical Analysis} \label{sec:sprm}

To motivate our \ffrob{} theoretical analysis, 
we review the theoretical analysis
for the simplified Probabilistic Roadmap (sPRM)~\citep{Kavraki98probabilisticroadmaps} over the class of robustly feasible motion planning problems. Two desirable properties for sampling-based motion planning algorithms are probabilistic completeness and exponential convergence.
Exponential convergence implies probabilistic completeness. 

\begin{defn}
An algorithm is {\em probabilistically complete} over a class of problems if and only if the probability that the algorithm halts and returns a solution is one in the limit as the number of time steps goes to infinity.
\end{defn}

\begin{defn}
An algorithm is {\em exponentially convergent} over a class of problems if and only if the probability that the algorithm has not terminated and returned a solution decreases exponentially in the number of time steps.
\end{defn}

The objective of a single-query motion planning problem is to find a collision-free trajectory in a $d$-dimensional configuration space ${\cal Q} \subseteq \mathbb{R}^d$ between a start configuration $q^0 \in {\cal Q}$ and a goal configuration $q^* \in {\cal Q}$. As typical in the analysis of motion planning algorithms, we restrict our analysis to Euclidean configuration spaces. We will call configurations and trajectories collision-free if all of their configurations are contained within ${\cal Q}$.  

\subsection{Robust Feasibility}

First, we identify a class of robustly feasible motion planning problems which have a nonzero volume of solutions. 
The robustness restriction is necessary because sampling-based algorithms have zero probability of generating samples on any particular lower dimensional sub-manifold of ${\cal Q}$. Having a nonzero volume of solutions ensures that the solutions are not at some segment constrained to such a sub-manifold. 

We give a definition of a robustly feasible motion planning problem that mixes the ideas of clearance~\citep{kavraki1998analysis} and $\epsilon$-goodness~\citep{kavraki1995randomized} in order to classify some motion planning problems where $q^0$ or $q^*$ is on the boundary of ${\cal Q}$ as robustly feasible. We will use $||x||$ as the Euclidean norm on points $x \in \mathbb{R}^d$. Let $\tau: [0, L] \to {\cal Q}$ be a trajectory of length $L$ from $q^0$ to $q^*$ such that $q^0 = \tau(0)$ to $q^* = \tau(L)$. Let $\chi(\tau; {\cal Q})$ give the {\em clearance} of $\tau$, the minimum distance from a configuration on $\tau$ to the boundary of ${\cal Q}$:

\begin{equation*}
\chi(\tau; {\cal Q}) = \inf_{t \in [0, L]} \inf_{x \in \partial {\cal Q}} ||\tau(t) - x||.
\end{equation*}

\begin{defn}
A motion planning problem is {\em robustly feasible} if there exists a trajectory $\tau$ from $q_+^0 \in {\cal Q}$ to $q_+^* \in {\cal Q}$ and $\delta > 0$ such that:
\end{defn}
\begin{itemize}
\item $\chi(\tau; {\cal Q}) \geq \delta$ and:
\item $\forall q \in {\cal Q}$ such that $||q - q_+^0|| \leq \delta/2$, \\
$(1 - t)q^0 + tq \in {\cal Q}$ and:
\item $\forall q' \in {\cal Q}$ such that $||q' - q_+^*|| \leq \delta/2$, \\
$(1 - t)q' + tq^* \in {\cal Q}$.
\end{itemize}

\begin{figure}
\includegraphics[width=0.49\textwidth]{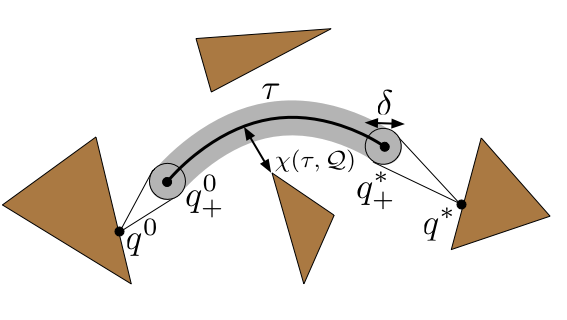}
\caption{A robustly feasible motion planning problem.} \label{fig:tube}
\end{figure}

This definition asserts that there is a trajectory with nonzero clearance such that the neighborhoods around its start and end configurations can "see" (admit a linear path between) $q^0$  and $q^*$ respectively. This implies that both $q^0$ and $q^*$ are $\epsilon$-good for some $\epsilon > 0$. This is a weaker condition than the assertion that there exists a direct trajectory between $q^0$ and $q^*$ with nonzero clearance. The latter assertion would disqualify motion planning problems where the start or goal are themselves on the boundary of ${\cal Q}$, even if a non-negligible volume of ${\cal Q}$ could see them. These kinds of motion planning problems are prevalent when grasping objects, making our definition useful when analyzing \prob{} problems. Figure~\ref{fig:tube} displays a motion planning problem that is robustly feasible under our definition of the term.

\subsection{Probabilistic Completeness}

This following theorem states that for any robustly feasible motion planning problem, there exists a finite sequence of $d$-spheres with nonzero volume for which the linear interpolation of any collection of configurations covering the spheres is a solution to the problem. Let $B(q, r)$ be the $d$-sphere centered at $q$ with radius $r$. 

\begin{thm} \label{sprm:balls}
For any robustly feasible motion planning problem, there exists a sequence of $k+1$, where $k = \lceil 2 L / \delta \rceil$, $d$-spheres $(B_0, B_1, ..., B_k)$ centered at $\tau(Li/k)$ for $i \in \{0, ..., k\}$ and each with radius $\delta/2$ such that any trajectory $\tau'$ linearly interpolated from $(q^0, q_0, q_1, ..., q_k, q^*)$, where $q_i \in B_i$ for $i \in \{0, ..., k\}$, is a collision-free solution to the motion planning problem.
\end{thm}

We give this proof, which is largely the same as a proof by~\cite{kavraki1998analysis}, in the appendix. 
This theorem directly reduces motion planning to a sampling problem by identifying these spheres. Thus, sampling-based algorithms, such at PRMs, can produce solutions to robustly feasible motion planning problems by sampling the space until generating samples within these spheres. 

Again, we consider completeness properties for the sPRM, a mathematically tractable variant of a PRM. The sPRM starts its roadmap with $V = \{q^0, q^*\}$ and, on each iteration, uniformly at random samples ${\cal Q}$ and connects each sampled configuration $q$ to all existing roadmap configurations $q'$ such that $\forall t \in [0, 1]$, $(1 - t)q + tq' \in {\cal Q}$, {\it i.e.} the linearly interpolated path between $q$ and $q'$ is collision-free.
Now we prove the main theorem, which is slightly modified from that of~\cite{kavraki1998analysis}. Let $\mu(Q)$ be the $d$-dimensional Lebesgue measure on $Q \subseteq {\cal Q}$.

\begin{thm} \label{sprm:complete}
The sPRM algorithm is probabilistically complete and exponentially convergent over the class of robustly feasible motion planning problems.
\begin{proof}

Consider any robustly feasible planning problem. By theorem~\ref{sprm:balls}, there exists a sequence of $k+1$, where 
$k = \lceil 2 L / \delta \rceil$, 
$d$-spheres with radius $\delta/2$ such that any collection of $k+1$ configuration samples that covers the spheres forms a collision-free, linearly interpolated trajectory. 
We will construct an upper bound on the probability that, after taking $n$ samples, sPRM has not covered all of the balls.
We will assume that the sPRM is able to directly sample ${\cal Q}$ by, for instance, sampling a hyperrectangle subset of $\mathbb{R}^d$ that contains ${\cal Q}$ and rejecting samples not contained ${\cal Q}$.

We begin by defining $\sigma$ to be the probability that a random sample is inside some particular sphere (and note that this probability is equal for all the spheres).
It is equal to the ratio of the measure of the sphere to the measure of the free configuration space $Q$:
\begin{equation}
\Pr[\text{sample is in } B_i] = \frac{\mu(B(\tau(Li/k), \delta/2))}{\mu({\cal Q})} = \sigma
\end{equation}

Note that $\sigma \in [0, 1]$ is a constant with respect to $n$. 
Now, the probability that all $n$ samples are outside ball $i$ is $(1-\sigma)^n$ which is bounded above by $e^{-\sigma n}$.   
The algorithm will fail if, for any ball, all $n$ samples are outside it; we bound this probability using the union bound:
\begin{align}
\Pr[\bigcup_{i=0}^k{\text{all } n \text{ samples are outside } B_i}] &\leq \sum_{i=0}^k e^{-\sigma n} \\
&= \Big\lceil \frac{2 L}{\delta} +1 \Big\rceil e^{-\sigma n}.
\end{align}

Note that $\lim_{n \to \infty} \lceil 2 L / \delta +1 \rceil e^{-\sigma n} = 0$. 
Because the probability of failure decreases is exponentially in $n$, the sPRM algorithm is exponentially convergent which implies it is probabilistically complete. \qed
\end{proof}
\end{thm}


\section{\ffrob{} Theoretical Analysis}  \label{sec:theory} 
\label{sec:analysis}

The probabilistic completeness and exponential convergence proofs for \ffrob{} build on the ideas from the sPRM proofs. The high level structure of the argument is the same. We first identify a class of robustly feasible \prob{} problems that have a nonzero volume of solutions. This is more complicated than in the pure motion planning case because we must reason not only about volumes of trajectories but also volumes of placements, grasps, and inverse kinematic solutions. Then, we describe a simplified version of \ffrob{}. Finally, we show that this version of \ffrob{} is probabilistically complete and exponentially convergent.


Because samples of configurations, poses, and grasps come from different domains, we need a definition of volume relative to each particular space. For the rest of the proof, let $\mu(E; X)$ generically be a measure on subsets $E$ of the set $X$ that assigns finite, nonzero measure to $X$. This could be for discrete $X$ a counting measure, for Euclidean manifolds a Lebesgue measure of appropriate dimensionality, for rotation groups a Haar measure, and so on. Furthermore, it will be useful to define a normalized version of this measure, $\tilde{\mu}$, where
\begin{equation*}
\tilde{\mu}(E; X) = \frac{\mu(E; X)}{\mu(X; X)}.
\end{equation*}

When picking and placing objects, we must frequently reason about the set of robot configurations that can reach a particular end-effector transformation. This set of inverse kinematics solutions is frequently a low dimensional sub-manifold in $\mathbb{R}^d$, which cannot always be sampled reliably. We denote the set of inverse kinematics solutions for a manipulator transform $t$ as:
\begin{equation*}
Q_{ik}(t) = \{q \in {\cal Q} \mid \proc{end-effector}(q) = t\}.
\end{equation*}

We assume that we can randomly sample from $Q_{ik}(t)$ with probability density bounded away from zero by some $\kappa > 0$.
This could be done by, for example, obtaining an analytical representation of the inverse kinematic solutions and sampling free parameters. 
This, particularly for high dimensional robots, is an open problem. 
The probabilistic completeness of \ffrob{} rests on having such a sampler.

\subsection{Robust Feasibility}

We now define a robustly feasible \prob{} problem by first introducing the components of \prob{} solutions that must themselves be robust for a \prob{} problem to be robust. 

\subsubsection{Mode-Constrained Motion Planning}

Many manipulation problems can be thought of as motion planning through multiple {\em modes}, where each mode $\omega$ defines constraints that restrict the system's operable state-space~\citep{HauserIJRR11}. Specifically, plans in \prob{} domains can be described as motion plans through an alternating sequence of transit and transfer modes~\citep{simeon2004manipulation}. A transit mode $\psi$ corresponds to the robot moving while not holding anything, and a transfer mode $\phi$ corresponds to the robot moving while holding an object at a fixed grasp. Each mode is defined by a set of parameters which give rise to its constraints.

\begin{defn}
A {\em transit mode} $\psi_i = \langle \kw{None}, \{p_i^{o_j} \mid \forall j \in [m]\} \rangle$ has parameters consisting of poses for each object $p_i^{o_j} \in {\cal P}^{o_j}$. The grasp is \kw{None}. A transit mode is legal if there are no collisions among the placed objects. A transit mode constrains the robot to move in $Q_{\psi_i} \subseteq {\cal Q}$, the subset of ${\cal Q}$ that does not collide with the placed objects.
\end{defn}

\begin{defn}
A {\em transfer mode} $\phi_i = \langle g_i^{o_{h_i}}, \{p_i^{o_j} \mid \forall j \in [m], j \neq h_i\} \rangle$ has parameters consisting of a grasp $g_i^{o_{h_i}} \in {\cal G}^{o_{h_i}}$ for an object $o_{h_i}$ and the poses for each other object $p_i^{o_j} \in {\cal P}^{o_j}$. A transfer mode is legal if there are no collisions among the placed objects. A transfer mode constrains the robot to move in $Q_{\phi_i} \subseteq {\cal Q}$, the subset of ${\cal Q}$ where neither it nor object $o_{h_i}$, held at grasp $g_i^{o_{h_i}}$, collide with the placed objects nor $o_{h_i}$ intersects with the robot.
\end{defn}

In both modes, the poses of the non-held objects are fixed and constrain legal movements of the robot. Additionally, in transfer modes, the held object must remain at the same grasp transform relative to the robot's end-effector, effectively altering the geometry of the robot. The state of the system can always be derived from just the current transit or transfer mode and the current robot configuration. As such, when the current mode is fixed, we can reason about the state of the system by just reasoning in the space of robot configurations subject to the mode.

The problem of planning robot movements subject to a mode is called a {\em mode-constrained} motion planning problem. This is simply a standard motion planning problem with an additional mode input $\omega$ which defines the {\em operable} state-space $Q_{\omega} \subseteq {\cal Q}$ of the problem. A mode-constrained motion planning problem is robustly feasible when the corresponding motion planning problem in the restricted configuration space $Q_{\omega}$ is robustly feasible.


Clearly, the sPRM is both probabilistically complete and exponentially convergent for robustly feasible mode-constrained motion planning problems by applying the exact same arguments from theorems~\ref{sprm:balls} and~\ref{sprm:complete}. 

\subsubsection{Multi-Mode-Constrained Motion Planning}

A robot must usually move in a sequence of modes in order to solve manipulation problems. Two modes are {\em adjacent} if the intersection between their operable system state-spaces is nonempty. Two unique transit modes cannot be adjacent because at least one object pose must differ between them, and the robot must grasp the object in order to move it between poses. Similarly, two unique transfer modes cannot be adjacent because an object pose or the grasp differs between them. In either case, the robot must change the current grasp to move the object or obtain a new grasp. 
Transit and transfer modes, however, can be adjacent. 

\begin{defn}
A transit mode $\psi_i$ and transfer mode $\phi_j$ are {\em adjacent} if and only if $p_{i}^{o_a} = p_{j}^{o_a}$ $\forall a \in [m], h_j \neq a$, and $Q_{\psi_i} \cap Q_{\phi_j} \neq \emptyset$. 
\end{defn}

In adjacent modes, the poses of the objects that are not grasped in the transfer mode are fixed between the two modes (reflecting that the robot can only manipulate a single object at a time) and the set of robot configurations that can move between the modes is nonempty. The system can perform a {\em mode switch} between two adjacent modes when it is at a state in the intersection of the two operable state-spaces. For transit and transfer modes, this is equivalent to a robot being at a configuration $q \in Q_{\psi_i} \cap Q_{\phi_j}$. Mode switches from $\psi_i$ to $\phi_j$ are \proc{Pick} actions. Conversely, mode switches from $\phi_j$ to $\psi_i$ are \proc{Place} actions. Both of these actions involve object $o_{h_j}$ and grasp $g_{j}^{o_{h_j}}$ from the transfer mode as well as pose $p_{i}^{o_{h_j}}$ from the transit mode. Precisely, the set of $\proc{Pick}$ actions $A_{\psi_i}^{\phi_j}$ that can switch from from $\psi_i$ to $\phi_j$ are
$$A_{\psi_i}^{\phi_j} = \{\proc{Pick}(o_{h_j}, p_{i}^{o_{h_j}}, g_{j}^{o_{h_j}}, q) \mid q \in Q_{\psi_i} \cap Q_{\phi_j}\},$$
and the set of $\proc{Place}$ actions $A_{\phi_j}^{\psi_i} $ that can switch from from $\phi_j$ to $\psi_i$ are
$$A_{\phi_j}^{\psi_i} = \{\proc{Place}(o_{h_j}, p_{i}^{o_{h_j}}, g_{j}^{o_{h_j}}, q) \mid q \in Q_{\phi_j} \cap Q_{\psi_i} \}.$$

Define $t(\psi_i, \phi_j) = p_{i}^{o_{h_j}} \times \proc{transform}(g_{j}^{o_{h_j}})^{-1}$ as the end-effector transform for grasping object $o_{h_j}$ at pose $p_{i}^{o_{h_j}}$ with grasp $g_{j}^{o_{h_j}}$. For notational simplicity, assume that the arguments to $t$ are unordered so $t(\phi_j, \psi_i) = t(\psi_i, \phi_j)$. $Q_{\psi_i} \cap Q_{\phi_j} \subseteq Q_{ik}(t(\psi_i, \phi_j))$ is the collision-free subset of the inverse kinematic solutions for the end-effector transform at the mode switch. These inverse kinematic configurations will serve as targets for motion planning to move between modes.

A sequence of $k$ legal, adjacent transit and transfer modes $(\omega_1, \omega_2, ..., \omega_k)$ is a {\em mode sequence}. A \prob{} mode sequence has $k-1$ mode switches; {\it i.e.}, $k-1$ \proc{Pick} and \proc{Place} actions. 
We will generically refer to the $i$th mode as $\omega_i$. Given both a mode sequence and whether $\omega_1$ is a transit or transfer mode allow us to determine the type of any other mode $\omega_i$ depending on whether $i$ is odd or even.

We will now look at {\em multi-mode-constrained} motion planning problems from a start configuration $q_0 \in Q_{\omega_1}$ to a set of end configurations $Q_* \subseteq Q_{\omega_k}$ through a fixed sequence of modes. These problems can be reduced to a sequence of $k$ mode-constrained motion planning problems. However, there is a complication that the start $q_i^0$ and goal $q_i^*$ for the $i$th mode-constrained motion planning problem (with the exception of the first and last problems) are not given. These must be chosen from the intersection of the neighboring modes such that $q_i^0 \in Q_{\omega_{i-1}} \cap Q_{\omega_i}$ and $q_i^* \in Q_{\omega_i} \cap Q_{\omega_{i+1}}$. The individual mode motion plans must connect continuously such that $q_i^* = q_{i-1}^0$. Thus, the problem requires choosing these target configurations as well as finding mode motion plans that connect between them. A multi-mode-constrained motion planning problem is robustly feasible when there exists a sequence of sets of target configurations with nonzero measure such that any mode-constrained motion planning problem between pairwise targets is robustly feasible.

\begin{defn}
A multi-mode-constrained motion planning problem is {\em robustly feasible} if for a mode sequence $(\omega_1, \omega_2, ..., \omega_k)$ there exists a sequence of sets of configurations $(Q_0, Q_1, ..., Q_k)$ and $\epsilon > 0$ such that:
\end{defn}
\begin{itemize}
\item $Q_0 = \{q_0\}$ and:
\item $\forall i \in [k-1]$ $Q_i \subseteq Q_{ik}(t(\omega_i, \omega_{i+1}))$ and $\tilde{\mu}(Q_i; Q_{ik}(t(\omega_i, \omega_{i+1}))) \geq \epsilon$ and:
\item $Q_k \subseteq Q_*$ and $\tilde{\mu}(Q_k; Q_*) \geq \epsilon$ and:
\item $\forall i \in [k]$ $\forall q_i^0 \in Q_{i-1}$ $\forall q_i^* \in Q_{i}$ the $\omega_i$ mode-constrained motion planning problem from $q_i^0$ to $q_i^*$ is robustly feasible.
\end{itemize}


\subsubsection{\prob{} Planning}

A multi-mode-constrained motion planning problem is easier than a \prob{} problem because the mode sequence is given as an input. 
We now return to full \prob{} problems that require selection of a mode sequence in addition to finding a multi-mode motion plan. 
The start mode must be $\omega_1  = \langle g_0^{o_{h_0}}, \{p_0^{o_1}, ..., p_0^{o_m}\} \rangle$. However, the goal mode can be any mode $\omega_k$ such that $g_k^{o_{h_k}} \in G_*^{o_{h_*}}$ and $\forall j \in [m]. p_k^{o_j} \in P_*^{o_j}$. 

Suppose now we must determine the sequence of modes. Starting from $\omega_1$ and for each newly chosen mode, each legal mode switch from the last mode $\omega_i$ to a new mode $\omega_{i+1}$ can be described by a single parameter. 
Between a transit mode $\omega_i = \psi_i$ and a new transfer mode, the object poses, with the exception of a grasped object, remain constant. Thus, the new transfer mode $\omega_{i+1} = \phi_{i+1}$ can be inferred from just $\psi_i$ and the specification of a free parameter $g_{i+1}^{o_{h_{i+1}}}$ for the resulting grasp. Similarly, between a transfer and transit mode, the object poses, with the exception of a grasped object, also remain constant leaving a free parameter $p_{i+1}^{o_{h_i}}$ for the new pose of $o_{h_i}$ to give the new mode. Thus, a sequence of modes starting from $\omega_1$ can be completely described by an alternating sequence of $(k-1)$ poses and grasps such as $(g_2^{o_{h_2}}, p_3^{o_{h_2}}, ..., p_k^{o_{k-1}})$. It is sufficient to choose these poses and grasps to identify a mode sequence. 

In order for a \prob{} problem to be robustly feasible, there must be a nonzero volume of mode sequences that admit robust multi-mode motion plans. As previously suggested, the set of length-$k$ mode sequences is contained in the space $\Theta$ formed by the Cartesian product of the $k-1$ alternating pose and grasp parameter domains ${\cal P}^{o_i}$ and ${\cal G}^{o_j}$. Suppose that we are considering mode sequences starting and ending with transit modes where the transfer modes interact with this prescribed sequence of objects $(o_{h_2}, o_{h_4}, ..., o_{h_{k-1}})$. The space containing the set of valid mode sequences is $\Theta = {\cal G}^{o_{h_2}} \times {\cal P}^{o_{h_2}} \times ... \times {\cal P}^{o_{h_{k-1}}}$. We can define a set of mode sequences $\theta = G_2^{o_{h_2}} \times P_3^{o_{h_2}} \times ... \times P_k^{o_{h_{k-1}}}$ by a set of values for each parameter such that $\theta \subseteq \Theta$. 

Let $P_i^{o_{h_{i-1}}}$ when $(i - j)$ is even and $G_j^{o_{h_j}}$ when $(i - j)$ is odd refer to these sets of grasps and poses that together define a collection of mode sequences from a prescribed sequence of transit modes and transfer modes. 
We could measure the volume of these mode sequences by just taking the product of each $\tilde{\mu}(P_i^{o_{h_{i-1}}}; {\cal P}^{o_{h_{i-1}}})$ and $\tilde{\mu}(G_j^{o_{h_j}}; {\cal G}^{o_{h_j}})$. However, the goal of these mode sequences is to reach a goal mode. This requires choosing grasps and poses within $G_*^{o_{h_*}}$ and $P_*^{o_i}$ for $i \in [m]$ respectively. A  \prob{} problem may specify a set of goal poses or grasps that is considerably smaller or has lower dimensionality than the full space of these values. For example, a problem could specify a goal set $P_*^{o_i} = \{p_*^{o_i}\}$ as a single pose, where clearly $\tilde{\mu}(P_*^{o_i} ; {\cal P}^{o_i}) = 0$. Yet, we still expect some of these problems to be robustly feasible because an algorithm could intentionally sample the goal set $P_*^{o_i}$ in addition to the full domain ${\cal P}^{o_i}$. Thus, we will define the measure of $P_i^{o_{h_{i-1}}}$ and $G_j^{o_{h_j}}$ as follows:
\[\tilde{\mu}(P_i^{o_{h_{i-1}}}) = \begin{cases} 
\tilde{\mu}(P_i^{o_{h_{i-1}}}; P_*^{o_{h_{i-1}}}) & P_i^{o_{h_{i-1}}} \subseteq P_*^{o_{h_{i-1}}}\\
\tilde{\mu}(P_i^{o_{h_{i-1}}}; {\cal P}^{o_{h_{i-1}}}) & \text{otherwise} 
\end{cases}\]
\[\tilde{\mu}(G_j^{o_{h_j}}) = \begin{cases} 
\tilde{\mu}(G_j^{o_{h_j}}; G_*^{o_{h_*}}) & o_{h_j} = o_{h_*} \text{ and } G_j^{o_{h_j}} \subseteq G_*^{o_{h_*}}\\
\tilde{\mu}(G_j^{o_{h_j}}; {\cal G}^{o_{h_j}}) & \text{otherwise} 
\end{cases}.\]

Intuitively, these measures are taken with respect to the set of goal values when the set to be measured is a subset of the goal values. Otherwise, the measures are taken with respect to the full domain of values. Finally, we arrive at the definition of a robustly feasible \prob{} problem.

\begin{defn}
A \prob{} problem $\Pi$ is {\em robustly feasible} if there exists a set of length-$k$ mode sequences $\theta = \{(\omega_1, \omega_2, ..., \omega_k), (\omega'_1, \omega'_2, ..., \omega'_k), ...\}$ with transfer modes involving a common sequence of objects $(..., o^*_{h_i}, o^*_{h_{i+2}}, ...)$ such that:
\end{defn}
\begin{itemize}
\item $\forall (\omega_1, \omega_2, ..., \omega_k) \in \theta$:
\begin{itemize}
\item $\omega_1  = \langle g_0^{o_{h_0}}, \{p_0^{o_1}, ..., p_0^{o_m}\} \rangle$.
\item $o_{h_i} = o^*_{h_i}$ if $\omega_i$ is a transfer mode.
\item $g_k^{o_{h_k}} \in G_*^{o_{h_*}}$ and $\forall j \in [m]. p_k^{o_j} \in P_*^{o_j}$.
\item The multi-mode motion planning problem from $q_0$ to $Q_*$ using $(\omega_1, \omega_2, ..., \omega_k)$ is robustly feasible.
\end{itemize}
\item For the $i$th set of transfer modes $\{\omega_i, \omega'_i, ... \}$, $\tilde{\mu}(P_i^{o_{h_{i-1}}}) > 0$.
\item For the $j$th set of transit modes $\{\omega_j, \omega'_j, ... \}$, $\tilde{\mu}(G_j^{o_{h_j}}) > 0$.
\end{itemize}

Intuitively, a robustly feasible \prob{} problem has non-negligible volumes of sequential poses and grasps such that any choice of these poses and grasps allows a sequence of robust motion plans that can connect them and solve the problem. The non-negligible volumes of poses relate to the robustness definition given by~\cite{van2009path}. Their definition says a problem is robust if there exists a sequence of poses that could be simultaneously perturbed by a small amount $\Delta$ without changing the feasibility of the resulting motion planning problems. Relating this to our definition, $\Delta$ is the radius of an $n$-sphere that is centered at each pose in the sequence where each $n$-sphere denotes a volume of safe poses.

Our robustness condition forbids overly constrained problems where solution mode sequences are restricted to a lower dimensional sub-manifold of the mode sequence parameter space. 
Such problems require special samplers that are able to produce tuples of values, possibly for different types of parameters, that satisfy a constraint among them. Specifying such a sampler for every combination of parameters is intractable because the number of combinations grows exponentially. 

\subsubsection{Example}

Consider a physical interpretation of robustness for placements. Suppose all modes are chosen from the set of mode sequences apart from the $i$th transfer mode. This consequently fixes the sequence of placements for all but the $i$th placement. The robustness condition asserts that for any combination of fixed placements, any choice of the $i$th placement from $P_i^{o_{h_{i-1}}}$, and therefore any choice of the $i$th transfer mode, will not collide with the existing placements. This allows for placements to be chosen independently with respect to each other. 

\begin{figure}
\centering
\includegraphics[width=0.35\textwidth]{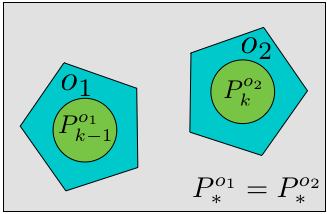}
\includegraphics[width=0.4\textwidth]{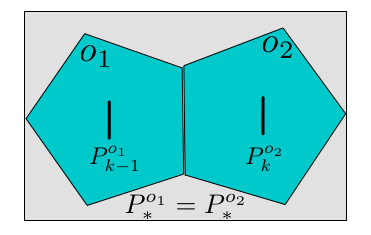}
\caption{Illustration of robust feasibility applied to placements. The goal poses $P_*^{o_1}$ and $P_*^{o_2}$ for the blue pentagons $o_1, o_2$ are poses contained in the grey region. The top scenario shows candidate $P_{k-1}^{o_1}$ and $P_k^{o_2}$ with nonzero areas. The bottom scenario only admits $P_*^{o_1}$ and $P_*^{o_2}$ that are lines giving each zero area.} \label{robust-poses}
\end{figure}

Figure~\ref{robust-poses} gives a example of a robust and non-robust problem involving goal poses. The top problem is robust because $P_{k-1}^{o_1}$ and $P_{k}^{o_2}$ are both balls of poses with nonzero volume meaning that $o_1$ and $o_2$ could be at any pair of poses from these sets respectively and satisfy the goal without collision. The bottom problem is not robust because $P_{k-1}^{o_1}$ and $P_{k}^{o_2}$ are lines indicating that the objects can only be moved up and down and result in a collision-free solution.

\subsection{Probabilistic Completeness}

Now that we have defined a robustly feasible \prob{} problem, we can return to analyzing \ffrob{}. Recall that on each iteration, \ffrob{} generates a finite number of samples of poses, grasps, and configurations and then performs a discrete search to determine if this set of samples is sufficient to solve the \prob{} problem. Upon the discrete search failing to find a plan, \ffrob{} increases the number of samples on the next iteration. The algorithm terminates if and when the discrete search finds a plan. For the analysis, we assume that samples $x$ are drawn from a space $X$ uniformly at random as denoted by $x \sim X$. 

\subsubsection{Convergence in Iterations}

Let $n$ represent the number of sampling iterations. Let $P^{o_i}(n)$, $G^{o_i}(n)$, $Q^{o_i}(n)$ be the sets of sampled poses, grasps, and inverse kinematic configurations involving object $o_i$ after $n$ sampling iterations. For $i \in [m]$, $P^{o_i}(0) = \{p_0^{o_i}\}$ and $G^{o_i}(0) = \emptyset$ if $p_0^{o_i} \neq \kw{None}$, otherwise, $P^{o_i}(0) = \emptyset$ and $G^{o_i}(0) = \{g_0^{o_i}\}$. However, $Q^{o_i}(0) = \emptyset$ for all $i \in [m]$. Additionally, let $(V(n), E(n))$ define the vertices and edges of the \crg{} after $n$ sampling iterations where $V(n) = \emptyset$ and $E(n) = \emptyset$.
On the $n$th sampling iteration, for all $i \in [m]$:
\begin{itemize}
\item $P^{o_i}(n) = P^{o_i}(n-1) \cup \{p^{o_i}, p_*^{o_i}\}$ where $p^{o_i} \sim {\cal P}^{o_i}$ and $p_*^{o_i} \sim P_*^{o_i}$.
\item $G^{o_i}(n) = G^{o_i}(n-1) \cup \{g^{o_i}, g_*^{o_i}\}$ where $g^{o_i} \sim {\cal G}^{o_i}$ and when $o_i = o_{h_*}$, $g_*^{o_i} \sim G_*^{o_{h_*}}$.
\item $Q^{o_i}(n) = Q^{o_i}(n-1) \cup \{q \sim Q_{ik}(p \times \proc{transform}(g)^{-1})  \mid p \in P^{o_i}(n),  g \in G^{o_i}(n)\}$. 
\end{itemize}

On each iteration and for each object, \ffrob{} samples a pose, goal pose, grasp, and goal grasp (if the object has a holding goal). Additionally, it samples an inverse kinematics solution for each existing pair of poses and grasps. Each pose and grasp pair will continue to have new inverse kinematic solutions sampled as $n$ increases. 
\ffrob{} will remain probabilistically complete and exponentially convergent as long as the number of samples on iteration $n$ is bounded by some polynomial in $n$.

To simplify the analysis, we will consider a simplified version of the \crg{} that is an sPRM in ${\cal Q}$. As stated in section~\ref{sec:sprm}, the sPRM is a roadmap that always attempts to connect every pair of sampled configurations, which leads to an exponentially convergent motion planning algorithm. We only construct one roadmap as opposed to growing a separate roadmap for each grasp or combination of object placements. Although grasping an object changes the geometry of the robot, it only decreases the operable ${\cal Q}$. The same is true for placing objects. Thus, the \crg{} grown in the full configuration space ${\cal Q}$ with the hand empty and no placed objects will be sufficient to also capture paths in different transit and transfer modes.

On each sampling iteration, the inverse kinematics solutions that were previously sampled are added to the \crg{}. 
A single free space robot configuration $q$ and a single goal robot configuration $q_*$ are also sampled and added to the \crg{}. 
Although in practice the \crg{} is constructed using larger sample batches, sampling individual configurations simplifies the analysis without loss of generality.
Then, the set of edges is constructed for every possible pair of vertices that do not have collisions:

\begin{itemize}
\item $V(n) = V(n-1) \cup Q^{o_1}(n) \cup ... \cup Q^{o_m} \cup \{q, q_*\}$ where $q \sim {\cal Q}$ and $q_* \sim Q_*$. 
\item $E(n) = \{(v, v'), (v', v) \mid v \in V, v' \in V, \proc{collision-free}(v, v'; {\cal Q}) \}.$
\end{itemize}

After the samples are selected for the $n$th iteration, the actions passed to the discrete planner are the following:

\begin{itemize}
\item Move actions: \\
$A_{move} = \{\proc{Move}(q \in V, q' \in V, (V, E)) \mid q \neq q' \}$.
\item Pick actions for $i \in [m]$: \\
$A_{pick}^i = \{\proc{Pick}(o_{i}, p \in P^{o_i}(n), g \in G^{o_i}(n), q \in Q^{o_i}(n)) \mid q \in Q_{ik}(p \times \proc{transform}.(g)^{-1})\}$
\item Place actions for $i \in [m]$: \\
 $A_{place}^i = \{\proc{Place}(o_{i}, p \in P^{o_i}(n), g \in G^{o_i}(n), q \in Q^{o_i}(n)) \mid q \in Q_{ik}(p \times \proc{transform}(g)^{-1})\}$.
\end{itemize}

Finally, because the heuristic search algorithms that work in practice do not improve symbolic planning's worst case complexity, we simplify \ffrob{} such that it performs an unguided search. 

We start by identifying the convergence rate as a function of the number of sampling iterations. We then perform a change of variables to show that it also converges exponentially in the number of time steps.

\begin{thm} \label{ffrob:sample}
For any robustly feasible \prob{} problem $\Pi$, the probability that \ffrob{} fails to find a solution decreases exponentially as the number of sampling iterations $n \to \infty$.
\begin{proof}

Recall that because \ffrob{}'s discrete search is complete, it will find a solution if a sufficient set of pose, grasp, and configuration samples are chosen. \ffrob{} will fail to find a solution by the $n$th sampling iteration if and only if the set of samples on the $n$th iteration does not contain a solution. We first divide this sampling failure into the disjunction of three disjoint failure events: an insufficient set of mode samples, an insufficient set of mode switch inverse kinematic samples assuming a sufficient set of mode samples, and an insufficient set of \crg{} configurations assuming sufficient sets of mode and mode switch samples. Let $N_{\ffrob{}}$ be a nonnegative random variable for the number of iterations before \ffrob{} finds a plan given a robustly feasible \prob{} problem. Let $N_{mode}$, $N_{switch}$, and $N_{\crg}$, be nonnegative random variables for the number of sampling iterations before \ffrob{} has produced a sufficient set of mode, mode switch, and \crg{} samples respectively. The probability that \ffrob{} fails to find a plan is the sum of the probabilities of the three disjoint failure events.
\begin{align*}
\Pr[&N_{\ffrob{}} > n] = \Pr[N_{mode} > n] \\
&+\Pr[N_{mode} \leq n, N_{switch} > n] \\
&+\Pr[N_{mode}, N_{switch} \leq n, N_{\crg{}} > n]
\end{align*}

We start with $\Pr[N_{mode} > n]$. Recall that a robustly feasible \prob{} problem has a non-negligible volume of mode sequences which can be represented by an alternating sequence of $(k-1)$ sets of pose and grasp samples. Let $P_i^{o_{h_{i-1}}}$ and $G_j^{o_{h_j}}$ be these sets of samples. The probability of a mode sampling failure is bounded by the sum of the probabilities of a pose and grasp sampling failure.
\begin{equation} \label{eq:1}
\Pr[N_{mode} > n] \leq \Pr[N_{pose} > n] + \Pr[N_{grasp} > n] 
\end{equation}

Because, on each iteration, \ffrob{} independently, uniformly at random samples a pose from both the full domain and the goal for each object, the probability that a new sample is in a set of poses $P_i^{o_{h_{i-1}}}$ is equal to the measure of $P_i^{o_{h_{i-1}}}$ using the relative measure $\tilde{\mu}$: 
\begin{equation*}
\Pr[p^{o_{h_{i-1}}} \in P_i^{o_{h_{i-1}}}] = \tilde{\mu}(P_i^{o_{h_{i-1}}}).
\end{equation*}

For simplicity, define $\rho_p = \min_i \Pr[p^{o_{h_{i-1}}} \in P_i^{o_{h_{i-1}}}]$. Using the union bound, the probability of a pose failure is less than the sum of the probabilities that all $n$ pose samples land outside $P^{o_{h_{i-1}}}(n) \cap P_i^{o_{h_{i-1}}}$.
\begin{align*}
\Pr[N_{pose} > n]  &= \Pr\Big[\bigvee_{i} (P^{o_{h_{i-1}}}(n) \cap P_i^{o_{h_{i-1}}} = \emptyset)\Big] \\
&\leq \sum_i \Pr\big[P^{o_{h_{i-1}}}(n) \cap P_i^{o_{h_{i-1}}} = \emptyset\big] \\
&\leq \sum_i \big(1- \Pr[p^{o_{h_{i-1}}} \in P_i^{o_{h_{i-1}}}]\big)^n   \\
&\leq (k-1)(1- \rho_p)^n \\
&\leq ke^{- \rho_p n} 
\end{align*}

Similarly, for a set of grasps $G_j^{o_{h_j}}$:
\begin{equation*}
\Pr[g^{o_{h_j}} \in G_j^{o_{h_j}}] = \tilde{\mu}(G_j^{o_{h_j}}).
\end{equation*}

Analogously, define $\rho_g = \min_i \Pr[g^{o_{h_j}} \in G_j^{o_{h_j}}]$ as the minimum probability of a pose or grasp from each set in the sequence. Then,

\begin{align*}
\Pr[N_{grasp} > n]  &= \Pr\Big[\bigvee_{j} (G^{o_{h_j}}(n) \cap G_j^{o_{h_j}} = \emptyset)\Big] \\
&\leq \sum_j \Pr\big[G^{o_{h_j}}(n) \cap G_j^{o_{h_j}} = \emptyset\big] \\
&\leq \sum_j \big(1- \Pr[g^{o_{h_j}} \in G_j^{o_{h_j}}]\big)^n   \\
&\leq (k-1)(1- \rho_g)^n  \\
&\leq ke^{- \rho_g n}.
\end{align*}

Moving on to $\Pr[N_{mode} \leq n, N_{switch} > n]$, recall that \ffrob{} samples an inverse kinematic configuration for all grasp and pose pairs on each iteration.
Thus, the events for obtaining successful mode samples and successful mode switch samples are not independent. Consider the following conditional probability $\Pr[N_{switch} > n \mid N_{mode} = i]$. Given $N_{mode} = i$, \ffrob{} will have $(n-i)$ iterations to generate the mode switches. To simplify the analysis, we upper bound $\Pr[N_{mode} \leq n, N_{switch} > n]$ by splitting the sampling into two failure cases, $[N_{mode} > n/2]$ and $[N_{mode} \leq n/2, N_{switch} > n \ ]$, which allows each case to be analyzed independently.
\begin{align*}
\Pr[N_{mode} \leq n, &N_{switch} > n]  \leq \Pr[N_{mode} > n/2] \nonumber\\
&+ \Pr[N_{mode} \leq n/2, N_{switch} > n]
\end{align*}




We already have a bound for $\Pr[N_{mode} > n/2]$ from equation~\ref{eq:1}. What remains is to bound $\Pr[N_{mode} \leq n/2, N_{switch} > n]$.
Each grasp and pose mode sample may be a seed for up to two inverse kinematic configurations. Without conditioning on the event that a full set of successful mode samples have been generated, it may be the case that what would be a successful inverse kinematics for a particular pose and grasp pair could still be a failure if it turned out that no full mode sequence using the pose and grasp was generated. Let $Q_i$ be the set of inverse kinematics solutions associated with the sampled mode sequence. 
Because the set of samples induces a robustly feasible multi-mode-constrained motion planning problem, there exists $\epsilon > 0$ such that $\tilde{\mu}(Q_i) \geq \epsilon$ $\forall i \in \{2, ..., k\}$. 
Recall that \ffrob{} samples inverse kinematic configurations with probability density bounded away from zero by $\kappa > 0$.
We will assume $\epsilon$ and $\kappa$ are the minimum volumes and densities respectively across any of the mode sequences in $\theta$.
Thus, $\Pr[q \in Q_i] \geq \kappa \epsilon$ $\forall i \in \{2, ..., k\}$. 
\begin{align}
\Pr&[N_{mode} \leq n/2, N_{switch} > n]  \nonumber\\
&= \sum_{i=1}^{n/2} \Pr[N_{mode} = i] \Pr[N_{switch} > n \mid N_{mode} = i] \nonumber\\
&\leq \Pr[N_{switch} > n \mid N_{mode} = n/2] \sum_{i=1}^{n/2} \Pr[N_{mode} = i] \nonumber\\
&\leq \Pr[N_{switch} > n \mid N_{mode} = n/2] \nonumber\\
&= \Pr\Big[\bigvee_{i=2}^{k} (Q(n) \cap Q_i = \emptyset)\Big] \nonumber\\
&\leq \sum_{i=2}^{k} \Pr\big[Q(n) \cap Q_i = \emptyset\big] \nonumber\\
&\leq \sum_{i=2}^{k} \big(1- \Pr[q \in Q_i]\big)^{n/2} \nonumber\\
&\leq (k-1)(1- \kappa\epsilon)^{n/2} \nonumber\\
&\leq k e^{-\kappa\epsilon n/2} \label{eq:2}
\end{align}


Finally, when analyzing $\Pr[N_{switch} \leq n, N_{\crg{}} > n]$, recall that the \crg{} samples a roadmap configuration per iteration, independently of any of the other created samples. Thus, we do not have the dependence complication we faced when sampling mode switches. We can upper bound the probability of the conjunction with the probability of the conditional: 
\begin{align*}
\Pr&[N_{mode} \leq n, N_{switch} \leq n, N_{\crg{}} > n] \\
&\leq \Pr[N_{\crg{}} > n \mid N_{mode}, N_{switch}].
\end{align*}

A successful set of mode switches will give rise to $(k-1)$ robustly feasible mode-constrained motion planning problems. Although the \crg{} does not know {\it a priori} which motion planning problems its samples will help solve, no part of its sampling is dependent on the targets because the samples are uniformly drawn from the unconstrained configuration space ${\cal Q}$. The samples created could be useful for any of the motion planning problems that arise. Let $L$ and $\delta > 0$ be the largest and smallest plan lengths and plan clearances across any of the robust motion planning problems. 
Using the union bound and theorem~\ref{sprm:complete},
\begin{equation} \label{eq:3}
\Pr[N_{\crg{}} > n \mid N_{mode}, N_{switch} \leq n] \leq k\Big\lceil \frac{2 L}{\delta} +1 \Big\rceil e^{-\sigma n}
\end{equation}

Combining our bounds on the different types of failures, according to equations \ref{eq:1}, \ref{eq:2}, and \ref{eq:3}, we have:

\begin{align*}
&\Pr[N_{\ffrob{}} > n]  \\
&\begin{aligned}
\leq & \Pr[N_{mode} > n/2] + \Pr[N_{switch} > n \mid N_{mode} = n/2] \\
&+ \Pr[N_{\crg{}} > n \mid N_{mode} \leq n, N_{switch} \leq n]
\end{aligned} \\
&\leq ke^{- \rho_p n/2} + ke^{- \rho_g n/2} + k e^{-\kappa\epsilon n/2} +  k \Big\lceil \frac{2 L}{\delta} +1 \Big\rceil e^{-\sigma n} \\
&= k\Big(e^{- \rho_p n/2} + e^{- \rho_g n/2} + e^{-\kappa\epsilon n/2} +  \Big\lceil \frac{2 L}{\delta} +1 \Big\rceil e^{-\sigma n} \Big).
\end{align*}

Thus, the probability that \ffrob{} fails decreases exponentially in $n$. \qed
\end{proof}
\end{thm}

\subsubsection{Convergence in Runtime}

While we have proven that \ffrob{} converges exponentially in the number of sampling iterations, this does not necessarily imply that it converges exponentially in runtime. For example, if the number of operations between each sampling iteration grows exponentially in $n$, an algorithm that converges exponentially in $n$ would only converge polynomially in the runtime $t$. For an example where $t = e^{n}$, $\Pr[\text{failure}] \leq e^{n} = e^{\ln{t}} = 1/t$.

Recall that symbolic planning is known to be PSPACE-Complete~\citep{bylander1994computational}. 
However, the discrete search will actually run in polynomial time of the number of samples because the addition of samples only increases the size of a variable's domain. 
The number of variables in the state-space $|{\cal V}|$ is fixed per problem based on the number of moveable objects $m$.
Thus, the size of the state-space grows only polynomially in the number of samples. 
The following lemma gives a very loose upper bound on the runtime of the discrete search on the $n$th sampling iteration.
 
\begin{lem} \label{ffrob:runtime}
The running time of the discrete search for a \prob{} problem with $m$ objects on the $n$th sampling iteration is $O(m^4n^{m+9})$.
\begin{proof}

Our symbolic planning problem can be viewed as a directed graph search problem where vertices $V'$ are possible states and edges $E'$ are actions performed from a specific state. A discrete search, such as Dijkstra's algorithm, can find the optimal solution to graph search problems in $O(|E'| + |V'| \log{|V'|})$. We compute $|V'|$ by analyzing the size of the state-space $S(n) = V'(n)$ after $n$ iterations. Given our sampling strategy, we have the following quantifies of samples after the $n$th iteration:
\begin{itemize}
\item Poses for object $o_i$: $|P^{o_i}(n)| = O(n)$
\item Grasps for object $o_i$: $|G^{o_i}(n)| = O(n)$
\item Robot inverse kinematic configurations: \\
$|Q(n)| \leq \sum_{i=1}^{m} \sum_{j=1}^{n} |P^{o_i}(j)||G^{o_i}(j)| = O(mn^3)$
\item Robot configurations: \\
$|V(n)| = O(n) + Q(n) = O(mn^3)$
\end{itemize}

In each state, the robot either has nothing in its hand with all the objects placed, or the robot is holding an object, so all but one objects are placed. Additionally, each state uses a single robot configuration. The size of $S$ is upper bounded by the combinations of samples for these two types of states.
\begin{align*}
|S(n)| &\leq |V(n)| \Big(\prod_{i = 1}^m |P^{o_i}(n)| + \sum_{i=1}^m |G^{o_i}(n)| \prod_{j \neq i }^m |P^{o_i}(n)|\Big) \\
&= O(mn^3(n^m + m(n \times n^{m-1}))) \\
&= O(m^2n^{m+3})
\end{align*}

Next, we compute $|E|$ by first calculating the number of possible actions given these samples.
\begin{itemize}
\item Move actions: $|A_{move}(n)| \leq |V(n)|^2 = O(m^2n^6)$
\item Pick actions: $|A_{pick}(n)| = |Q(n)| = O(mn^3)$
\item Place actions: $|A_{place}(n)| = |Q(n)| = O(mn^3)$
\end{itemize}
\begin{align*}
|A(n)| &= |A_{move}(n)| +  |A_{pick}(n)| +  |A_{place}(n)| \\
&= O(m^2n^6).
\end{align*}

From $|S(n)|$ and $|A(n)|$, we can obtain a loose upper bound on the number of edges in the state-space graph:
\begin{align*}
|E'(n)|  &= |S(n)| |A(n)| \\
&= O(m^4n^{m+9}).
\end{align*}

The number of edges dominates the Dijkstra runtime, thus the discrete search runs in $O(m^4n^{m+9})$. \qed
\end{proof}
\end{lem}

This analysis was meant to prove that the discrete search runs in polynomial time in $n$. The polynomial bound is loose and is not indicative of the runtime of common \prob{} instances, particularly when the search guided by a heuristic. Moreover, in practice, sampling is often more expensive than searching due to the geometric overhead from collision checks and inverse kinematics.

\begin{thm} \label{ffrob:combined}
\ffrob{} is probabilistically complete and exponentially convergent.
\begin{proof}

We can compute the change of variables between iterations $n$ and runtime $t$ using lemma~\ref{ffrob:runtime}. Each iteration takes $O(m) + O(m) + O(mn^2) = O(mn^2)$ time steps to produce the new samples before running the discrete search. This is clearly dominated by the planning time.
\begin{align*}
t(n) &= \sum_{i=1}^{n}O(m^4i^{m+9}) \\
&= m^4 \sum_{t=1}^{n}O(i^{m+9}) \\
&= O(m^4n^{m+10}) \\
&\leq C m^4n^{m+10} \text{ for some } C \text{ and } \forall n \geq n_0
\end{align*}

Inverting this mapping gives:
\begin{equation*}
n(t) \geq  \Big(\frac{n}{Cm^4}\Big)^{1/(m+10)}.
\end{equation*}

Let $C = \min(\rho_p/2, \rho_g/2, \kappa\epsilon/2, \sigma)$ and $T_{\ffrob{}}$ be a nonnegative random variable for the number time steps before \ffrob{} finds a solution.
\begin{align*}
\Pr[T_{\ffrob{}} > t] &\leq  k \Big\lceil 2 L/\delta + 4 \Big\rceil e^{-C (\frac{n}{Km^4})^{1/(m+10)}} \\
&= O\Big(e^{-n^{1/(m+10)}}\Big)
\end{align*}
Again, $m$ is fixed per the problem which slows but does not stop the exponential convergence. \qed
\end{proof}
\end{thm}

\subsubsection{Corollaries} 

We arrive at the following corollaries.

\begin{cor}
\ffrob{} has both a finite expected runtime and finite variance in runtime.

\begin{proof}
This follows from the result that an exponentially convergent algorithm has a finite expected runtime and finite variance in runtime~\citep{HauserLatombe}. \qed
\end{proof}
\end{cor}

If \ffrob{} is modified such that it continues to increase its set of samples even after finding a solution, it will converge to a solution that is in the minimal length robust set of mode sequences. This means it will find a solution that uses the fewest \proc{Pick} and \proc{Place} mode switches out of the set of solutions that are in a robust set of mode sequences. Moreover, this convergence will also be exponential.

\begin{cor} ~\label{cor:optimal}
The probability that \ffrob{} has not identified a solution contained in the minimal length robust set of mode sequences decreases exponentially in the number of time steps.
\begin{proof}
By theorem~\ref{ffrob:combined}, the probability that \ffrob{} will not have samples corresponding to a solution for any robust set of mode sequences decreases exponentially as $t \to \infty$. On each iteration, Dijkstra's algorithm will find the optimal plan in terms of the number of \proc{Pick} and \proc{Place} actions (we will let \proc{Move} actions have a cost of zero while \proc{Pick} and \proc{Place} have unit cost). As soon as samples that admit a minimal length mode sequence have been generated, Dijkstra's algorithm will continue to produce a plan with this mode sequence length forever because additional samples could only produce a solution with a smaller length. \qed
\end{proof}
\end{cor}

\ffrob{} will not terminate or declare when it has found an optimal solution, but it with high probability will eventually find a solution with minimal mode-sequence length in finite expected runtime. Additionally, the discrete search could be another optimal search algorithm instead of Dijkstra, such as A* with an admissible heuristic, that may have better practical performance. In symbolic planning, a cost sensitive version of $h_{max}$ is guaranteed to be admissible. As an additional practical aside, once a candidate solution is found, the discrete search need not visit states which have summed path cost and admissible heuristic cost that exceeds the cost of the previously optimal plan. While this also does not affect the theoretical analysis, it will prune the search space in practice.

\section{Experiments}\label{sec:exp}

We experimentally evaluated seven configurations of \ffrob{} using eight problems spanning rearrangement planning, NAMO, nonmonotonic planning, and task and motion planning.

\subsection{Problems}

\begin{figure*}
     \centering
     \includegraphics[width=0.44\textwidth]{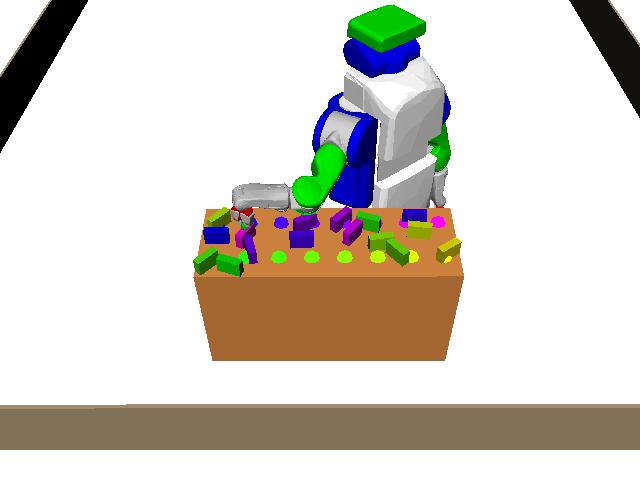}
     \hspace{0.1\textwidth}
     \includegraphics[width=0.44\textwidth]{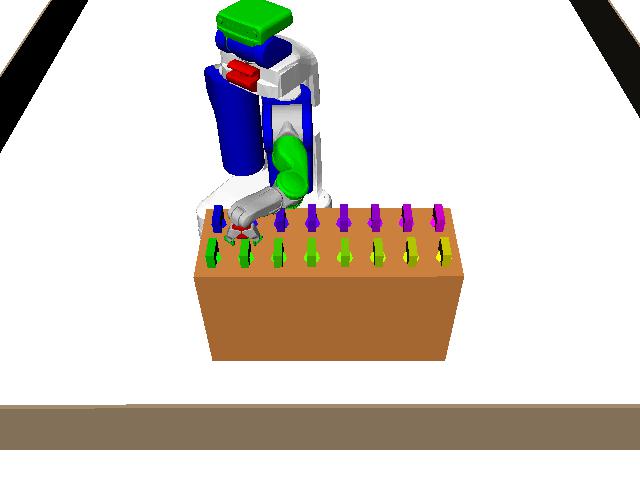}
     \caption{The second state and last state on a plan for problem 1-2.}
     \label{problem:grid}
\end{figure*}

Problems 1-1 \& 1-2 are simple rearrangement problems inspired by~\cite{krontirisRSS2015}. Each block has a specified goal configuration as represented by the color gradient.
The robot may use a single top grasp. 
Problem 1-1 has two rows of four blocks. Problem 2-1 has two rows of eight blocks and is displayed in figure~\ref{problem:grid}.

\begin{figure*}
     \centering
     \includegraphics[width=0.44\textwidth]{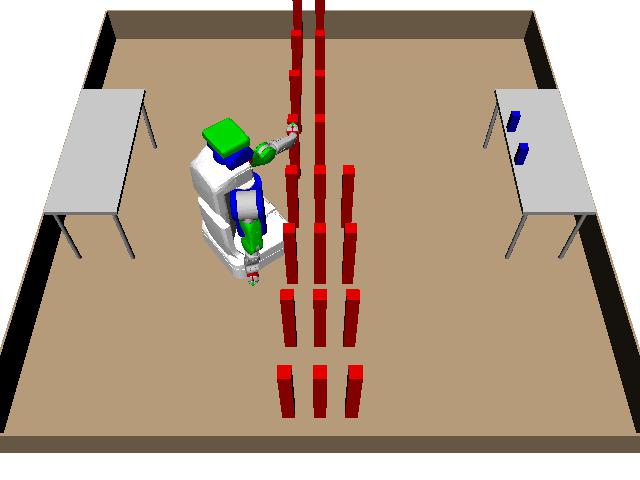}
     \hspace{0.1\textwidth}
     \includegraphics[width=0.44\textwidth]{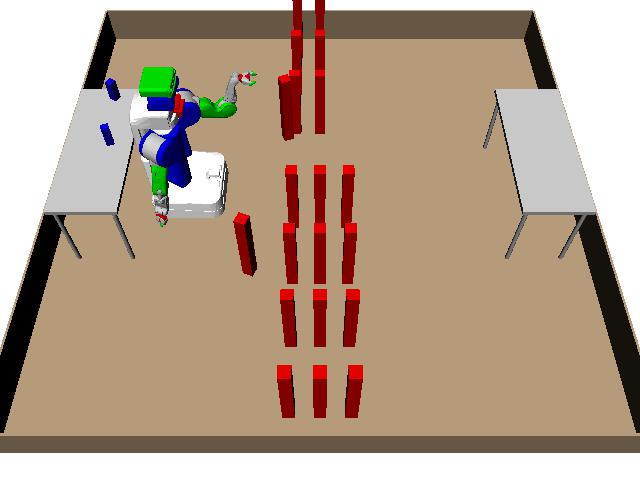}
     \caption{The second state and last state on a plan for problem 2-2.}
     \label{problem:namo}
\end{figure*}

Problems 2-1 \& 2-2 combine NAMO with tabletop pick-and-place. They require the robot to move the two blue blocks from the right table to the left table and return to its initial configuration. In order to reach the blue blocks, the robot must first move several red pillars out of way to clear a path for its base. The wall of pillars is composed of two segments where the top segment is thinner than the bottom segment. This is designed to test whether the heuristic can identify actions that lead to shorter plans that only move the top pillars. Problem 2-1 has a top segment of width one and a bottom segment of width two. Problem 2-2, displayed in figure~\ref{problem:namo}, has a top segment of width two and a bottom segment of width three.
 
\begin{figure*}
     \centering
     \includegraphics[width=0.44\textwidth]{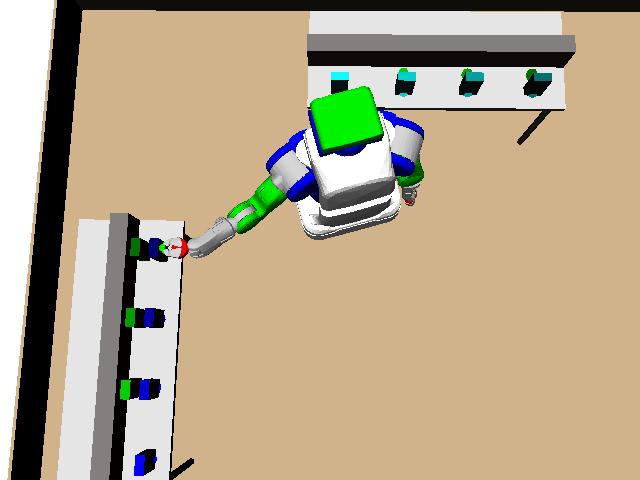}
     \hspace{0.1\textwidth}
     \includegraphics[width=0.44\textwidth]{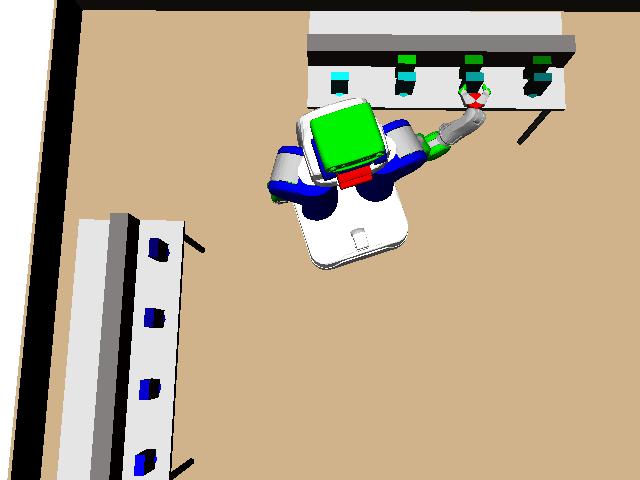}
     \caption{The second state and last state on a plan for problem 3-2.}
     \label{problem:nonmonotonic}
\end{figure*}

Problems 3-1 \& 3-2 are examples of highly nonmonotonic problems, problems that require undoing goal conditions along solutions. The robot must move the green blocks from the left table to a corresponding position on the right table. Both the initial and goal poses are blocked by four blue and cyan blocks respectively. Critically, the blue and cyan blocks have goal conditions to remain in their initial poses. This is the source of the nonmonotonicity as the robot must undo several goals by moving the blue and cyan blocks in order to solve the problem. Problem 3-1 includes only one green block. Problem 3-2, displayed in figure~\ref{problem:nonmonotonic}, includes three green blocks.

\begin{figure*}
     \centering
     \includegraphics[width=0.44\textwidth]{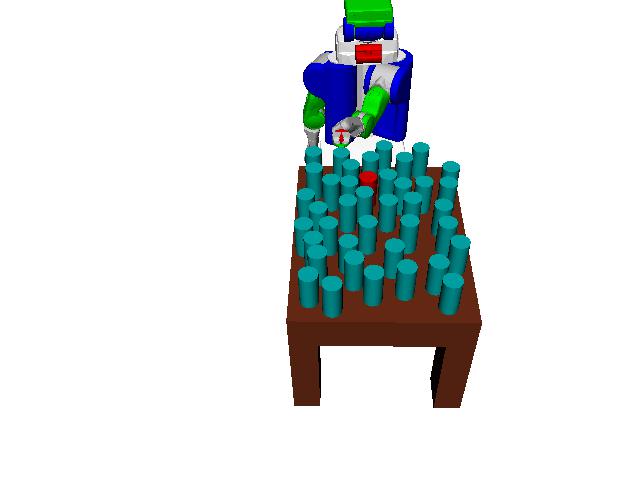}
     \hspace{0.1\textwidth}
     \includegraphics[width=0.44\textwidth]{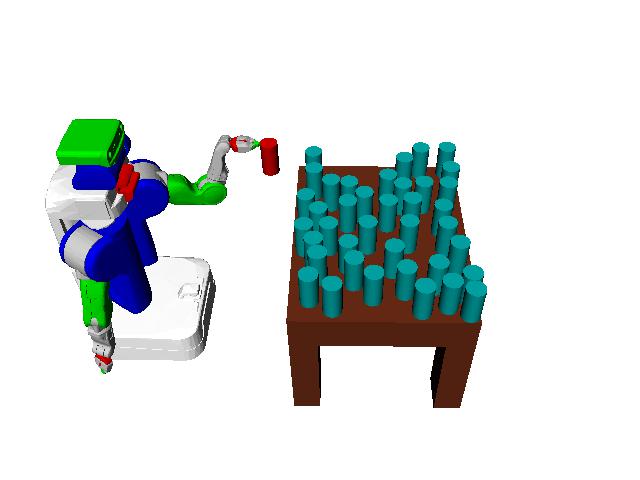}
     \caption{The second state and last state on a plan for problem 4.}
     \label{problem:srivastava}
\end{figure*}

Problem 4 in figure~\ref{problem:srivastava} is from~\cite{Srivastava14}. The robot must retrieve the red cylinder from within the cluttered table of cyan cylinders. The goal conditions are for the robot to be holding the red cylinder at its initial configuration.

\begin{figure*}
     \centering
     \includegraphics[width=0.44\textwidth]{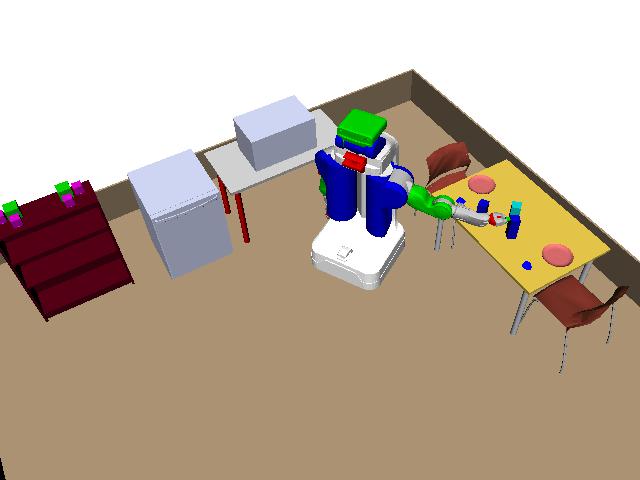}
     \hspace{0.1\textwidth}
     \includegraphics[width=0.44\textwidth]{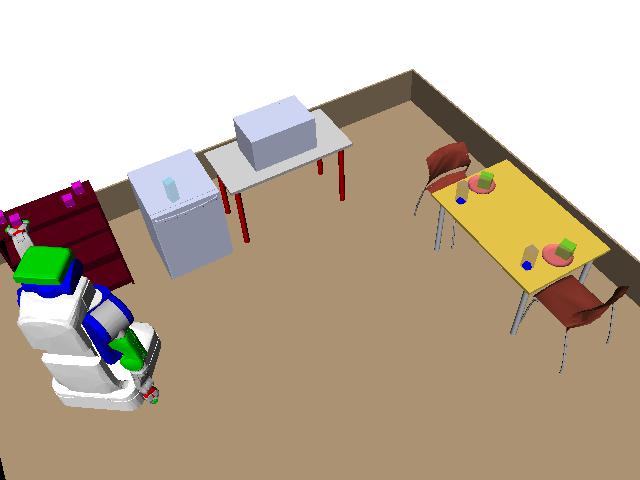}
     \caption{The second state and last state on a plan for problem 5.}
     \label{problem:dinner}
\end{figure*}

Problem 5 in figure~\ref{problem:dinner} is a task and motion planning problem in a cooking domain. The robot must retrieve two heads of cabbage (green blocks) from the shelves, clean the heads, cook the heads, and place them on the plates. Turnips (pink blocks) are present on the shelves and must be moved to reach the cabbage. However, the turnips must be returned to the shelves if moved. The robot must also wash the cups (blue and cyan blocks) and set the table using the blue cups. The cyan cup is not needed for dinner. This problem requires the purely symbolic literals \kw{Cleaned} and \kw{Cooked} as well as actions \proc{Clean} and \proc{Cook} which are shown in figure~\ref{fig:additional}. The top of the dishwasher is used to clean food and cups. The top of the microwave is used to cook food. Objects become transparent when cleaned or cooked.

\subsection{Heuristics}

We experimented with several versions of \ffrob{} using different heuristics.
For all heuristics, as previously described, we automatically generate new samples if the \ffrob{} heuristic is infinite before planning because a finite heuristic value is a necessarily condition for feasibility. The~following heuristics are compared in the experiments:

\begin{enumerate}
\item \proc{H}$_0$: The heuristic returns 0.
\item \proc{H}$_{Goals}$: This calls $\proc{unsatisfied-goals}$ to 
 return the number of unsatisfied goal conditions.
\item \proc{H}$_{FF}$: This is the original FF heuristic based only on simple conditions. 
It explicitly ignores complex conditions, namely reachability conditions.
This heuristic represents the semantic attachments strategy of~\cite{dornhege09icaps}.
\item \proc{H}$_{MaxRob}$: This uses \proc{max-comb} when performing \proc{compute-costs} to produce its estimate.
\item \proc{H}$_{AddRob}$: This uses \proc{add-comb} when performing \proc{compute-costs} to produce its estimate.
\item \proc{H}$_{FFRob}$: This finds a relaxed plan using the $\proc{add-comb}$ version of \proc{compute-costs}
and $\proc{extract-relaxed-plan}$.
\item \proc{H}$_{FFRob}$, \proc{HA}: This finds a relaxed plan using the $h_{add}$ version of \proc{compute-costs}
and $\proc{extract-relaxed-plan}$ and computes first goal-achieving helpful actions based on that plan.
\end{enumerate}

We use deferred best-first search, described in appendix~\ref{sec:deferred}, as the search control for our experiments. Deferred best-first search typically outperforms standard best-first search 
because of its lazy evaluation of successors. 

\subsection{Implementation}

We implemented \ffrob{} in Python using the OpenRAVE robotics framework~\citep{openrave} for simulation.
We use a simulated PR2 robot with a mobile base and a single active manipulator. 
Thus, there are 10 active degrees of freedom.
We used Open Dynamics Engine~\citep{smith2005open} for collision checks.
Additionally, we employ bounding boxes and cache collision checks in order to reduce the collision-checking.

The sampling parameters $\theta$ used in our experiments where chosen relatively arbitrarily and are fixed across all problems and heuristics.
In practice, we do not increase the sampling parameter sizes upon a sampling failure. 
We will restrict the robot to four side grasps per objects except on problems 1-1 \& 1-2 where we use a single top grasp. 
We randomly sample 25 general poses per object type in addition to 5 poses per specified symbolic region. We bias the general poses to be collision-free in the start state.
We attempt to sample a single grasp trajectory per grasp and object pose. 
We enforce timeouts of 30 iterations for \proc{s-pick-place} due to inverse reachability, inverse kinematics, or motion planning failures. We increase this number to 50 for objects that have an explicit goal condition.
We reuse placement trajectories for objects with similar geometries.
We impose a pruning rule that dynamically limits the number of successor \proc{Place} actions considered that do not achieve a goal to 5. This allows a large number of placements to be created for constrained problems without greatly increasing the branching factor.

The approach trajectories sampled using $\proc{s-appr-traj}$ control only the robot manipulator, and the roadmap trajectories sampled using $\proc{s-traj}$ control only the robot base.
We use a star-graph \crg{} for all problems except problems 2-1 \& 2-2 where moveable objects are located on the floor. For these problems the \crg{} is fixed degree-PRM with a degree of 4 and 50 sampled intermediate roadmap configurations.

\subsection{Results}


\begin{table*}
\centering
\begin{tabular}{
||c||
g|c||
g|c||
g|c||
g|c||
g|c||
g|c||}
\hline
P &\multicolumn{2}{|c||}{\proc{H}$_{Goals}$}&\multicolumn{2}{|c||}{\proc{H}$_{FF}$}&\multicolumn{2}{|c||}{\proc{H}$_{MaxRob}$}&
\multicolumn{2}{|c||}{\proc{H}$_{AddRob}$}&\multicolumn{2}{|c||}{\proc{H}$_{FFRob}$}&\multicolumn{2}{|c||}{\proc{H}$_{FFRob},\proc{HA}$}\\
\hline
&
 \% & sec &
 \% & sec &
 \% & sec &
 \% & sec &
 \% & sec &
 \% & sec  \\
\hline
1-1&
100 & 5 &
100 & 2 &
0 & - &
100 & 4 &
100 & 4 &
100 & 2 \\
\hline
1-2&
98 & 27 &
96 & 35 &
0 & - &
98 & 26 &
100 & 44 &
100 & 23 \\
\hline
2-1&
40 & 78 &
80 & 73 &
94 & 74 &
98 & 34 &
94 & 33 &
96 & 26 \\
\hline
2-2&
0 & - &
12 & 182 &
46 & 115 &
62 & 91 &
82 & 60 &
76 & 50 \\
\hline
3-1&
20 & 245 &
92 & 109 &
0 & - &
90 & 104 &
88 & 57 &
96 & 19 \\
\hline
3-2&
0 & - &
0 & - &
0 & - &
0 & - &
0 & - &
72 & 135 \\
\hline
4&
0 & - &
0 & - &
30 & 97 &
46 & 126 &
78 & 38 &
74 & 31 \\
\hline
5&
0 & - &
0 & - &
0 & - &
74 & 56 &
84 & 137 &
76 & 44 \\
\hline
\end{tabular}
\caption{Experiment results over 50 trials.} \label{table:ffrob}
\end{table*}

\begin{figure}
     \centering
     \includegraphics[width=0.49\textwidth]{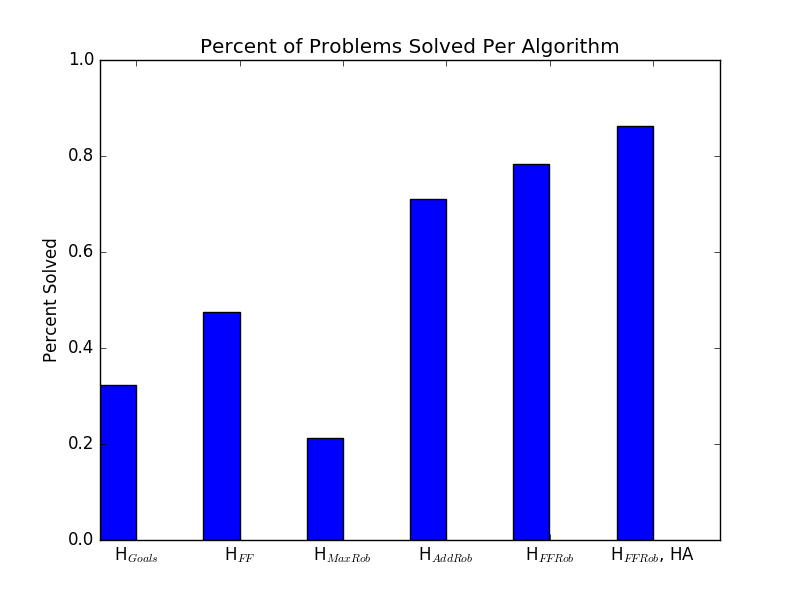}
     \caption{A bar graph of the overall success rate across all problems per algorithm.}
     \label{fig:percent_solved}
\end{figure}

Table~\ref{table:ffrob} displays the results of the experiments.
Each problem and algorithm combination was simulated over 50 trials. Each simulation had a timeout of 300 seconds. 
Simulations were performed on a single Intel Xeon E5 v3 2.5GHz processor.
Each entry in the table reports {\em success rate} (\%) and mean of the {\em runtimes} (sec) in seconds for solved instances. 
The runtimes measure the full algorithm runtime, including the sampling, collision-checking, pre-processing, and discrete planning.
We observed that the data often has large runtime outliers causing the mean runtime to be larger than the median runtime.
Runtime entries with a dash (-) indicate that the algorithm did not solve any of the simulations for that problem.
Figure~\ref{fig:percent_solved} displays the overall percent of solved instances across all problems for each algorithm.
\proc{H}$_0$ was unable to solve any of the problems and is excluded from the results.

\proc{H}$_{FFRob}, \proc{HA}$ gave the best performance in both success rate and runtime. 
Helpful actions improved the performance of \proc{H}$_{FFRob}, \proc{HA}$ over \proc{H}$_{FFRob}$.
\proc{H}$_{AddRob}$ performed worse than \proc{H}$_{Goals}$ and \proc{H}$_{FF}$ although it was not strictly worse across all problem instances.
In particular, \proc{H}$_{AddRob}$ performed poorly on problems 1-1, 1-2, and 3-1 because it requires about the same amount of overhead to evaluate 
as \proc{H}$_{FFRob}$ while providing a much less informative heuristic estimate.
\proc{H}$_{FF}$ performed worse than \proc{H}$_{FFRob}$ indicating that reachability information is vital for producing successor heuristic for geometrically non-trivial task and motion planning problems.

Problem 1-2 can be to compared the {\em grid @ tabletop (14 objects)} problem of~\cite{krontirisRSS2015}. 
All but 1 of our algorithms were able to solve it with an above 95 percent success ratio in less than 40 seconds. 
The fact that \proc{H}$_{Goals}$ was able to solve problems 1-1 and 1-2 indicates that they require little heuristic guidance. Because each object has an explicit goal, \proc{H}$_{Goals}$ corresponds well with the actual distance to the goal, which is approximately twice \proc{H}$_{Goals}$. In the event that the \proc{H}$_{Goals}$ must move an object more than once along the plan, it will rely on brute force search to reach a lower heuristic value. Thus, we observed that \proc{H}$_{Goals}$ required more expansions to solve a problem than \proc{H}$_{FFRob}$ but was able to perform each expansion more quickly because of the lower heuristic overhead. 

Problems 2-1 \& 2-2 demonstrate that \ffrob{} can solve both NAMO and traditional pick-and-place problems using the same algorithm.
Problems 3-1 \& 3-2 proved the most difficult because of their large amount of nonmonotonicity. Even in discrete settings, nonmonotonicity causes FF to produce poor heuristic estimates because of its delete-relaxation. However, our best algorithm manages to solve problem 3-2 demonstrating the ability to tackle problems with these elements.
The results of problem 4 suggest that strong heuristics are particularly necessary in problems with large branching factors resulting from large numbers of movable objects. 
Problem 5 demonstrates that \ffrob{} is able to quickly solve a long-horizon, real-world problem involving symbolic actions, cluttered environments, and nonmonotonic requirements. 

For the problems we posed, we noticed that \ffrob{} never performs more than one iteration of sampling and planning. 
Verifying that the discretized planning problem does not have a solution is computationally expensive because the worst case complexity grows exponentially in the number of moveable objects. In practice, terminating the search after a finite amount of time to generate new samples will result in better performance. Generally, if the discrete search using \proc{H}$_{FFRob}, \proc{HA}$ is given a feasible set of samples, it will identify a solution before a short timeout for non-adversarial problems. 
Thus, it can be advantageous to restart with a new set of samples even in cases where the original set of samples would have been sufficient.


\section{Conclusion}

\ffrob{} is a probabilistically complete algorithm for task and motion planning. It uses a EAS, a generalized action representation, to model discretized planning problems with complex conditions. We adapt the FF heuristic to EAS problems in order to provide strong heuristic guidance for these problems. \ffrob{} iteratively discretizes a task and motion planning problem and runs an EAS planner to evaluate if the current set of samples is sufficient for a solution. This leads to a probabilistically complete algorithm as \ffrob{} will, with probability approaching one, generate a set of samples that contains a solution.
In our results, we show that including geometric information in a heuristic is critical for the resulting search algorithm to efficiently solve manipulation problems involving interesting geometric constraints.
Additionally, we show that a single algorithm can efficiently solve a diverse set of task and motion planning problems.

Future work includes analytically and empirically investigating the quality of solutions returned by \ffrob{} with respect to costs. In corollary~\ref{cor:optimal}, we briefly remark that \ffrob{}, with an optimal implementation \proc{search}, will produce a solution no longer than the minimum length robustly feasible solution in finite time. This is the setting where \proc{Pick} and \proc{Place} actions have unit costs while \proc{Move} actions have zero cost. This can be easily extended to problems in which \proc{Pick}, \proc{Place}, \proc{Move}, and other actions have arbitrary, nonnegative fixed costs. A more interesting case is when \proc{Move} actions have costs dependent on the control effort required to move between their start and end configuration. This requires identifying a class of {\em robustly optimal} \prob{} problems and investigating whether sampling-based algorithms are {\em asymptotically optimal}, {\it i.e.} whether they with high probability converge to an optimal solution~\cite{KFIJRR11}. We suspect that \ffrob{} would be asymptotically optimal for a similar set of conditions as those presented in section~\ref{sec:analysis}.

Because \ffrob{} samples continuous values independently of its search, it often produces many unnecessary samples. 
This is undesirable because sampling can be time intensive and a large number of samples can slow \proc{search}.
For example, in figure~\ref{problem:srivastava}, \ffrob{} samples poses, grasps, configurations, and trajectories for each cylinder on the table although only a few are required to reach the red cylinder. Future work involves using the planning to guide the sampling such as done by~\cite{GarrettIROS15}.
Additional future work involves applying \ffrob{} to different manipulation tasks or generally planning domains involving continuous variables. For domains involving stacking, where there are many possible stacking combinations, the combinatorial growth in possible samples may overwhelm \ffrob{}. In which case, a careful sampling strategy would be needed.



\section{Acknowledgements}
We gratefully acknowledge support from NSF grants 1122374, 1420927, and 1523767, from ONR grant N00014-14-1-0486, and from ARO grant W911NF1410433.  Any opinions, findings, and conclusions or recommendations expressed in this material are those of the authors and do not necessarily reflect the views of our sponsors.

\newpage
\appendix

\section{Search Appendix} \label{search-appendix}

\subsection{Best-First Search}

Best-first search extracts the element in the queue that minimizes a cost function $f(n)$. 
Common cost functions include the path cost $f(n) \equiv n.g$ (Dijkstra's algorithm), the sum of path cost and heuristic cost ($A^*$), a weighted sum of path cost and heuristic cost $f(n; w) \equiv (1-w)n.g + (w)n.h$ (weighted $A^*$), and the heuristic cost $f(n) \equiv n.h$ (greedy best-first).

\begin{figure}[h!]
\begin{codebox}
\Procname{$\proc{bfs-extract}(Q, f)$}
\li \kw{return} $\proc{pop-min}(Q, f)$
\end{codebox}

\begin{codebox}
\Procname{$\proc{bfs-process}(Q, s', n;  \proc{h})$}
\li $\proc{push}(Q, \proc{StateN}(s', n.g + 1, \proc{h}(s'), n))$
\end{codebox}
\caption{Best-first search extract and process procedures.}
\end{figure}

\subsection{Deferred Best-First Search} \label{sec:deferred}

Deferred best-first search (also called lazy greedy search) is a variant of standard best-first search~\citep{helmert2006fast}.
It deferring the evaluation of successors states until they are extracted from the queue in order to reduce the number of heuristic evaluations.
Successors states temporarily use their parent's heuristic cost while in the queue. 
The intuition behind this approach is that successors of a state $s$ are often added to the queue in an order given by helpful actions such that states believed to be closer to the goal are processed first.
If a extracted state $s'$ has a low heuristic cost, its own successors will temporarily have that cost when they are added to the queue. 
Thus, the full set of successors of $s$ will likely not be processed because the search greedily proceeds down the subtree rooted at $s'$. 
The \proc{extract} procedure is the same for standard best-first search but the \proc{process} procedure is slightly modified to use the parent state's heuristic cost. 

\begin{figure}[h!]
\begin{codebox}
\Procname{$\proc{dbfs-process}(Q, s', n;  \proc{h})$}
\li $\proc{push}(Q, \proc{StateN}(s', n.g + 1, \proc{h}(\underline{n.s}), n))$
\end{codebox}
\caption{Deferred best-first search process procedure.}
\end{figure}

\section{Review of sPRM Analysis Appendix}

\subsection{Proof of Theorem 1}

\begin{thm} \label{balls}
For any robustly feasible motion planning problem, there exists a sequence of $k+1$, where $k = \Big\lceil \frac{2 L}{\delta} \Big\rceil$, $d$-spheres $(B_0, B_1, ..., B_k)$ centered at $\tau(Li/k)$ for $i \in \{0, ..., k\}$, each with radius $\delta/2$, such that any trajectory $\tau'$ linearly interpolated from $(q^0, q_0, q_1, ..., q_k, q^*)$, where $q_i \in B_i$ for $i \in \{0, ..., k\}$, is a collision-free solution to the motion planning problem.
\begin{proof}

First consider the following lemma.

\begin{lem} \label{subsets}
$B_{i+1} \subseteq B(\tau(Li/k), \delta)$
\begin{proof}

By our construction, using $x \leq \lceil x \rceil$,
\begin{align}
||\tau(L(i+1)/k) - \tau(Li/k)|| &\leq L(i+1)/k - Li/k \\
&= L/k \\
&\leq L/(2L/\delta) \\
&= \delta/2.
\end{align}

Now for any $q_{i+1} \in B_{i+1}$, by the triangle inequality,
\begin{align}
||q_{i+1} - \tau(Li/k)|| &\leq ||q_{i+1} - \tau(L(i+1)/k)|| \\
&\;\;\;\;+ ||\tau(L(i+1)/k) - \tau(Li/k)|| \nonumber \\
&\leq \delta/2 + \delta/2 \\
&= \delta.
\end{align}

Thus, each $q_{i+1}$ is contained in $B(\tau(Li/k), \delta)$ which implies $B_{i+1} \subseteq B(\tau(Li/k), \delta)$. \qed
\end{proof}
\end{lem}

Because $\tau$ has at least $\delta$ clearance from obstacles, all points within $B(\tau(t), \delta)$ for $t \in [0, L]$ are collision-free. Moreover, line segments between any two points in $B(\tau(t), \delta)$ for $t \in [0, L]$ are collision-free because because they are contained in $B(\tau(t), \delta)$ by convexity of the $d$-sphere. Using lemma~\ref{subsets}, the line segment between any $q_{i} \in B_i$ and any $q_{i+1} \in B_{i+1}$ is collision-free because both $q_i$ and $q_{i+1}$ are contained within $B(\tau(Li/k), \delta)$. By applying this to all $i, i+1$, we see that the linearly interpolated trajectory $(q_0, q_1, q_2, ..., q_k)$ is thus collision-free. Finally, each segment from $q^0$ to $q_0 \in B_0$ or from $q^*$ to $q_k \in B_k$ is collision-free by the problem being robustly feasible. So, the linearly interpolated trajectory $(q^0, q_0, q_1, q_2, ..., q_k, q^*)$ is also collision-free. \qed

\end{proof}
\end{thm}



\end{document}